\documentclass[]{elsarticle}

\usepackage{multirow}
\usepackage{hyperref}
\usepackage{amsmath}
\usepackage{caption}
\usepackage{color}
\usepackage{amsmath,amssymb}
\usepackage{subcaption}

\usepackage{changes}

\usepackage[utf8]{inputenc}
\usepackage[T1]{fontenc}

\newcommand{\KE}[1]{{{\color{cyan}{Ke: #1}}}} 
\newcommand{\MATHIAS}[1]{{{\color{red}{Mathias: #1}}}}  
\newcommand{\NORBERT}[1]{{{\color{blue}{Norbert: #1}}}}

\begin{document}

\begin{frontmatter}

\title{Photometric Multi-View Mesh Refinement for High-Resolution Satellite Images}

\author[ethz]{Mathias Rothermel\corref{mycorrespondingauthor}} \author[us]{Ke Gong} \author[us]{Dieter Fritsch} \author[ethz]{Konrad Schindler} \author[us]{\;\;\;Norbert Haala}

\cortext[mycorrespondingauthor]{Corresponding author}

\address[us]{Institute for Photogrammetry, University of Stuttgart}
\address[ethz]{Photogrammetry and Remote Sensing, ETH Zurich}

\begin{abstract}
 Modern high-resolution satellite sensors collect optical imagery with ground sampling distances (GSDs) of 30-50cm, which has sparked a renewed interest in photogrammetric 3D surface reconstruction from satellite data. State-of-the-art reconstruction methods typically generate 2.5D elevation data. Here, we present an approach to recover full 3D surface meshes from multi-view satellite imagery. The proposed method takes as input a coarse initial mesh and refines it by iteratively updating all vertex positions to maximise the photo-consistency between images. Photo-consistency is measured in image space, by transferring texture from one image to another via the surface. We derive the equations to propagate changes in texture similarity through the rational function model (RFM), often also referred to as rational polynomial coefficient (RPC) model. Furthermore, we devise a hierarchical scheme to optimise the surface with gradient descent.  In experiments with two different datasets, we show that the refinement improves the initial digital elevation models (DEMs) generated with conventional dense image matching. Moreover, we demonstrate that our method is able to reconstruct true 3D geometry, such as facade structures, if off-nadir views are available.
 \end{abstract}


\end{frontmatter}

\section{Introduction}
Automatic reconstruction of surface models from airborne and space-borne imagery is a long-standing  problem in photogrammetry and computer vision. For imagery collected by airborne platforms, state-of-the-art algorithms allow for country-scale reconstructions with an impressive level of detail and a high degree of robustness, as demonstrated for instance by the city models now included in virtual online globes.
In many applications built-up areas are most interesting and thus mapped on a regular basis, often with centimeter resolution. The most common product are still 2.5D digital surface models (DSMs) generated from airborne nadir imagery. More recent camera systems can also collect oblique views and enable the reconstruction of true 3D structures, in particular facade elements.
Compared to airborne imagery, traditional satellite sensors acquired images with lower resolution, less redundancy, and only nadir views. Consequently, 2.5D DSMs are still the predominant representation for satellite-based reconstructions.
The launch of new satellites with steerable high-resolution sensors, such as for instance WordView3, has led to a renewed interest in detailed reconstruction from spaceborne imagery. With these new systems, datasets with down to 0.3m GSD and high redundancy can be collected, see for example recent benchmarks \cite{Bosch2016,avibench}. Given such data, the extraction of real 3D geometry, like balconies and other facade structures, seems to be in reach. Most existing algorithms, however, only produce 2.5D height maps or surfaces \cite{Kuschk2013,Wohlfeil2012,dAngelo2011,shean2016automated,Gong2019} and do not even attempt to recover 3D details.

Besides the 2.5D scene representation, conventional reconstruction pipelines have further drawbacks. They are often based on pair-wise (dense) stereo and subsequent fusion of stereo models, which does not fully exploit the multi-view redundancy.
Moreover, the dominant binocular stereo algorithm in those pipelines is Semi-Global Matching (SGM) \cite{Hirschmuller2008}, due to its good compromise between quality and computational cost. The price to pay is that SGM and its variants are, by construction, subject to modeling errors such as fronto-parallel bias (caused by rectangular matching windows) and a preference for areas of constant disparity \cite{roth2019reduction,scharstein2017semi}. It is also well-documented that methods that estimate sub-pixel disparities in discrete disparity space introduce further systematic errors \cite{shimizu2002precise,szeliski2004sampling,gehrig2007improving}.
The fusion of individual stereo models into a single, consistent height field is most often done with heuristic rules, which certainly improve robustness and accuracy, but are nevertheless sub-optimal. In particular, visibility and occlusions are often handled poorly, or not at all.

To sidestep the mentioned limitations, we propose a novel reconstruction approach that uses a 3D mesh representation. Our method is a local optimisation starting from an initial mesh, i.e., it refines an existing surface model, for instance a conventional 2.5D stereo result.
Following \cite{delaunoy2008minimizing,vu2012high}, we assume that the coarse, initial mesh is topologically correct and we refine it by iteratively moving its vertices in the direction that most reduces the texture transfer error across all views.
Technically, this is implemented as a variational energy minimisation, subject to a surface smoothness prior. Here, we formulate the corresponding energy function for the RPC model, the dominant sensor model for satellite images. We demonstrate, for the first time, the reconstruction of full 3D surface structure, by incorporating satellite views with large off-nadir angles. Moreover, we show that the refinement also tends to improve the accuracy of the 2.5D elevation values.

\section{Related Work}

The majority of reconstruction algorithms for optical satellite imagery employ the scheme of pairwise epipolar rectification and dense stereo matching, followed by the fusion of depth maps  \cite{Kuschk2013,Wohlfeil2012,dAngelo2011,shean2016automated,Gong2019}. Epipolar rectification warps an image pair such that corresponding pixels share the same row index. This reduces the correspondences search space to 1D and accelerates pixel-wise matching. In contrast to pinhole cameras, the object-to-image space mapping for satellite sensors is usually described with the rational function model (RFM), also known as rational polynomial coefficients (RPC) model \cite{kratky1989line,tao2001comprehensive,dial2002block,kim2006comparison,poli2012review,hartley1997cubic}. Since standard epipolar geometry is not valid for this projection model, rectification algorithms like \cite{fusiello2000compact,loop1999computing,pollefeys1999simple} cannot be applied. However, \cite{Kim2000} have shown that, locally, correspondences are located on a pair of epipolar curves across the images. This finding enables rectification of complete satellite images by resampling the original images along the epipolar curves \cite{Wang2010, Wang2011, Oh2011}. Due to computational efficiency and low memory consumption, many reconstruction pipelines employ some variant of SGM for dense stereo matching. The classical SGM is implemented in \cite{dAngelo2011} and they show state-of-the-art performance on the ISPRS benchmark \cite{reinartz2010benchmarking}. \cite{d2016improving} investigate the compensation of \textit{overcounting} \cite{drory2014semi} in the context of SGM for satellite imagery and observe improved density but decreased precision. A hierarchical version of SGM is employed by \cite{Gong2019}, \cite{rothermel2012sure} to limit the disparity search range and reduce memory footprint and computation time, while also reducing matching ambiguities. \cite{DeFranchis2014} and the top-ranked competitors showed state-of-the-art performance on the IARPA satellite benchmark \cite{Bosch2016} with off-the-shelf SGM implementations from open source libraries \cite{opencv_library} and with NASA's open stereo reconstruction software \cite{shean2016automated}. Beside vanilla SGM, the latter also implements the More Global Matching \cite{facciolo2015mgm}, using a modified cost aggregation scheme that aims for globally more consistent disparities. In order to avoid costly energy minimisation altogether, \cite{Wang2017} construct dense correspondence maps from a sparse set of tie points, via edge aware interpolation.

To obtain a consistent representation of the surface, individual stereo models have to eventually be fused . This is typically realized during DSM generation, by binning 3D points from multiple models into a planimetric 2D grid, followed by various filtering strategies to derive a single elevation value per grid cell -- accomplished often by simple median filtering \cite{Kuschk2013,dAngelo2012,Wang2017}. \cite{Facciolo2017} account for changing surface modes (e.g, vegetation) caused by different acquisition dates. They propose $k$-median clustering for each grid cell and favour observations from lower clusters. In the spirit of bilateral filtering, \cite{Qin2017} apply median filtering on heights of multiple neighbouring grid cells that share the same intensities. \cite{kuschk2017spatially} cast the fusion of multiple stereo models or elevation maps as a convex energy minimization problem and solve it with a primal-dual algorithm, including an additional planarity prior in the form of a TV-L1 and TGV-L1 penalty.

Some approaches circumvent the somewhat cumbersome pairwise processing and subsequent fusion. Notably, early approaches of dense surface refinement were published in \cite{helava1988object,wrobel1987facets}.  Rasterized surface elevations and the corresponding object-space appearance are simultaneously estimated using iterative non-linear least squares. Thereby the surface elevations are refined such that gray or RGB values across multiple views (linked by the elevation values) are consistent. Beside the 2.5D representation, and thus the inability to reconstruct 3D surfaces, those early approaches lack proper occlusion handling or any form of geometric regularization.
 \cite{dAngelo2012} directly estimate a DSM  by assigning photometric similarity costs to a regular 3D cost structure in object space. The final elevation map is derived by semi global optimization. However, the authors find only limited  gains compared to the more prevalent late fusion of binocular stereo models. Similarly, \cite{Wang2016} estimate an elevation grid and additionally fuse semantic information from a set of satellite images and corresponding semantic segmentation maps. To estimate surface elevations, PatchMatch Belief Propagation \cite{besse2013patchmatch} is employed to maximize an energy function that encourages consistency of appearance and semantics across several images.
Additionally, smoothness of semantics, height values and surface orientations are enforced. \cite{Pollard2007,Pollard2010} propose a probabilistic voxel-based model to jointly reconstruct surface voxels and their corresponding colours. To our knowledge, this is the only published method in the satellite domain which is capable of extracting real 3D geometry. It does, however, not include any explicit surface prior, and \cite{Ozcanli2015} found that both urban and rural reconstructions are less accurate than those from pairwise matching and late fusion. 

Much less research exists for the satellite domain when compared to the conventional pinhole camera model, presumably because of the limited availability of high-resolution data.
We thus review  relevant work on mesh refinement in the close range and airborne domains.   %
Typical approaches start by reconstructing depth maps, or point clouds with normals, using Multi View Stereo (MVS), for example  \cite{schoenberger2016mvs,Furu:2010:PMVS,goesele2007multi,galliani2016gipuma}. Next, they employ a volumetric approach to bootstrap a topologically correct mesh representation (e.g., \cite{kazhdan2013screened,jancosek2011multi,ummenhofer2015global,labatut2009robust,zach2007globally,fuhrmann2014floating}). For the latest results we refer the interested reader to one of the active MVS and reconstruction benchmarks \cite{seitz2006comparison,schoeps2017cvpr,Knapitsch2017}. In order to recover fine details and improve precision, the vertex positions of such surface meshes can be further refined  \cite{delaunoy2008minimizing,vu2012high}. An energy composed of the (multi-view) texture transfer error and a smoothness prior is minimised with gradient descent. \cite{li2016efficient} accelerate the process by limiting the refinement to regions that feature geometric variances. An alternate minimisation of the reprojection error and mesh denoising is given by \cite{li2015detail}. A guideline for the mesh refinement by semantic information is pursued in \cite{blaha2017semantically,romanoni2017multi}: semantic consistency across views is enforced and smoothness priors for each class are adapted individually.
To the best of our knowledge, all refinement algorithms were formulated for pinhole  camera models and cannot be directly applied to satellite data. %

\section{Methodology}
\newcommand{\bfS}{S}
\newcommand{\bfI}{\mathcal{I}}
\newcommand{\bfPi}{\mathbf{\Pi}}
\newcommand{\bfx}{\mathbf{x}}
\newcommand{\bfd}{\mathbf{d}}
\newcommand{\bfX}{\mathbf{X}}
\newcommand{\bfn}{\mathbf{n}}
\newcommand{\bfN}{\mathbf{N}}
\newcommand{\bfD}{\mathbf{D}}
\newcommand{\bfP}{\mathbf{P}}
\newcommand{\bfO}{\mathbf{O}}
 The basic idea of photometric mesh refinement is to position the vertices of a mesh surface, such that texture transferred from an image $i$ to any other image $j$ via the surface should match the original texture of the target image $j$. As long as the textures are not in correspondence, they give rise to a gradient, which can be propagated through the sensor model to obtain a gradient per vertex, defining the direction in which it should be displaced to increase the texture similarity.
 Iterative gradient decent yields a refined mesh with maximal similarity, respectively minimal photometric reprojection error. For clarity, we first review the computation of gradients for the pinhole camera model in section \ref{sec:review}, before extending it to the RFM in section \ref{sec:rpccrefinement}. Additionally, section \ref{sec:impl_details} discusses implementation details.

\subsection{Mesh Refinement with Frame Sensors} \label{sec:review}
As a background for the subsequent adaption to satellite imagery we review the computation of gradients induced by photometric (dis-)similarities under the pinhole model. Let $\mathcal{S}$ denote the (infinite) set of admissible 2D surface manifolds in $\mathbb{R}^3$. 
The overall photoconsistency $\mathcal{M}$ is composed of individual terms $\mathcal{M}_{ij}: \mathcal{S} \rightarrow \mathbb{R}^1$ that measure the photoconsistency between image $\bfI_j: \Omega_j \rightarrow  \mathbb{R}^1$ to image $\bfI_i: \Omega_i \rightarrow \mathbb{R}^1$, when projected onto each other via the surface $\bfS$ $\in$  $\mathcal{S}$:
\begin{equation}
\mathcal{M}(\bfS)=\sum_i\sum_{j\neq i}{\mathcal{M}_{ij}(\bfS)}. 
\end{equation}
Here $\Omega_j$ and $\Omega_i$ are the image regions in $I_j$ and $I_i$ that see the same surface area $\Omega$ on $\bfS$. Let $\Pi_{j},\Pi_{i}: \mathbb{R}^3 \rightarrow \rm I\!{R}^2$ be the projections that map object coordinates to image coordinates of $\bfI_j$, respectively $\bfI_i$, and let $\Pi^{-1}_{i, \bfS}$ and $\Pi^{-1}_{j, \bfS}$ denote the re-projection from the respective image to the surface $\bfS$. The transfer function relating image coordinates in the two views is given by $\bfI_{j}\circ\Pi{j}\circ\Pi^{-1}_{i, \mathcal{S}}$, such that the pairwise photoconsistency becomes 
\begin{equation}
\mathcal{M}_{ij}(\bfS)=\int_{\Omega_i} { M(\bfI_i,\bfI_j \circ \Pi_j \circ \Pi^{-1}_{i, \bfS})} d\mathbf{x}_i,
\end{equation}
where $M$ is a measure of photo-consistency. In our case we seek to minimize the negative zero-normalised cross-correlation
$M(\mathbf{a},\mathbf{b})=-{ZNCC(\mathbf{a},\mathbf{b})}$.
Using the chain rule, the variation of $\mathcal{M}_{ij}$ with respect to an infinitesimal variation of the surface is given by
\begin{equation} \label{equ:before}
\frac{\partial\mathcal{M}_{ij}(\bfS+\epsilon\delta\bfS)}{\partial\epsilon}
\Bigg\vert_{\epsilon=0}=\int_{\Omega_i}\partial_2M(\mathbf{x}_i)D\bfI_j(\mathbf{x}_j)D\Pi_j(\mathbf{X})
\frac{\partial\Pi^{-1}_{i,\mathcal{S}+\epsilon\delta\mathcal{S}}(\mathbf{x}_i)}{\partial\epsilon}
\Bigg\vert_{\epsilon=0} d\mathbf{x}_i\;,
\end{equation}
where $\partial_2M(\mathbf{x}_i)$ denotes the derivative of the similarity measure w.r.t.\ the second argument $\bfI_j$. $D\bfI_j$ is the gradient in image $\bfI_j$ and $D\Pi_j$ is the derivative of the object-to-image space mapping w.r.t.\ an object point on the surface. Let $\mathbf{d}$ be the ray from the projection center of view $i$ through pixel coordinate $\mathbf{x}_i$. The term
\begin{equation}
\frac{\partial\Pi^{-1}_{i,\mathcal{S}+\epsilon\delta\mathcal{S}}(\mathbf{x}_i)}{\partial\epsilon}
\Bigg\vert_{\epsilon=0}
\end{equation}
represents the change along $\mathbf{d}$ when moving the surface by $\delta S$.  
\begin{figure}
\begin{subfigure}{0.5\textwidth}
\includegraphics[width=\linewidth]{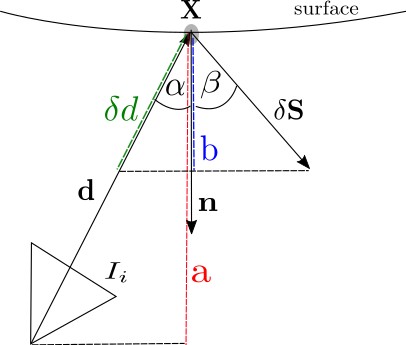}
\caption{} \label{fig:1a}
\end{subfigure}
\hspace*{\fill} %
\begin{subfigure}{0.4\textwidth}
\includegraphics[width=\linewidth]{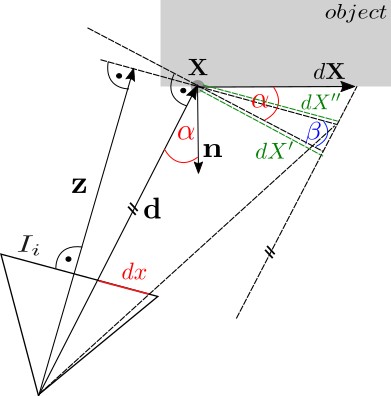}
\caption{} \label{fig:1b}
\end{subfigure}
\caption{(a): Variation of $\mathbf{d}$ induced by the surface variation $\mathbf{\delta S}$. Let $-\mathbf{n}^T\mathbf{d}=|\mathbf{d}|\cos{\alpha}:=a$ and $\mathbf{n}^T\delta\mathbf{\mathcal{S}}=|\delta \mathcal{S}|\cos{\beta}:=b$. The intercept theorem yields $\delta d=|\mathbf{d}| b/a$.
(b): Relation of displacement within the surface vs.\ displacement in the image. %
From $\cos{\alpha}=dX'/dX=-\mathbf{n}^T\mathbf{d}/|\mathbf{d}|$ and $\cos{\beta}=dX'/dX''=z/|\mathbf{d}|$ follows $dX''=-dX\mathbf{n}^T\mathbf{d}/z$. The intercept theorem yields $dx=-dX\mathbf{n}^T\mathbf{d}/z^{2}$
}\label{fig:1}
\end{figure}
This change can be computed geometrically using trigonometric functions and the intercept theorem (see figure \ref{fig:1a}), which leads to 
\begin{equation} \label{equ:deltad}
\frac{\partial\Pi^{-1}_{i,\mathcal{S}+\epsilon\delta\mathcal{S}}(\mathbf{x}_i)}{\partial\epsilon}
\Bigg\vert_{\epsilon=0}=\frac{\mathbf{n}^T\delta S(\mathbf{X})}{\mathbf{n}^T \mathbf{d}_i} \mathbf{d}_i
\end{equation}
with the surface normal $\mathbf{n}$. The change in image coordinates caused by an infinitesimal displacement of the object coordinates (see Figure \ref{fig:1b}) 
is computed as
\begin{equation} \label{equ:deltaS}
d\mathbf{x}_i=-\frac{\mathbf{n}^{T}\mathbf{d}_i d \mathbf{X}}{z^2}.
\end{equation}
Here $z$ represents the z-component of a surface point in the camera coordinate system of view $i$. By substituting (\ref{equ:deltad}) in (\ref{equ:before}), and using (\ref{equ:deltaS}) to change the integration domain from image space to the surface,  we obtain
\begin{equation} \label{equ:final}
\frac{\partial\mathcal{M}_{ij}(\mathcal{S}+\epsilon\delta\mathcal{S})}{\partial\epsilon}
h\Bigg\vert_{\epsilon=0}=-\int_{\Omega_{\mathcal{S}}}\partial_2M(\mathbf{x}_i)D\bfI_j(\mathbf{x}_j)D\Pi_j(\mathbf{X}) \frac{\mathbf{d}_i}{z^2} \mathbf{n}^T \delta \mathcal{S}d\mathbf{X}.
\end{equation}
It has been shown \cite{delaunoy2011gradient,solem2005geometric} that for the variation $\delta \mathcal{M}$ of the photo-consistency and the variation $\delta \mathcal{S}$ of the surface, the gradient vector field $\nabla \mathcal{M}$ fulfills
\begin{equation} \label{equ:ext1}
\delta \mathcal{M}=
\frac{\partial\mathcal{M}_{ij}(\mathcal{S}+\epsilon\delta\mathcal{S})}{\partial\epsilon}
\Bigg\vert_{\epsilon=0}=\int_{\Omega_{\mathcal{S}}}\nabla \mathcal{M}_{ij}(S) \delta\mathcal{S}d\mathbf{X},
\end{equation}
and the gradient descent flow is given by $-\nabla \mathcal{M}(\mathcal{S})$.
Consequently, by comparing (\ref{equ:final}) and (\ref{equ:ext1}), the gradient of the matching function is
\begin{equation}
\nabla \mathcal{M}_{ij}(S)=-\phi_{\Omega_{\mathcal{S}}}\left[\partial_2M(\mathbf{x}_i)D\bfI_j(\mathbf{x}_j)D\Pi_j(\mathbf{X})\frac{\mathbf{d}_i}{z^2}\right]\mathbf{n}.
\end{equation}
The flag $\phi_{\Omega_{\mathcal{S}}}$ accounts for visibility, evaluating to 1 if the surface is visible in both views $\bfI_i,\bfI_j$, and to 0 otherwise. Note that this continuous formulation can be directly used to compute gradients of discrete surfaces, for example triangular meshes. In practice a translation of each single vertex is computed by weighted integration of gradients over its one-ring neighborhood of triangles. For more details the interested reader is referred to \cite{delaunoy2011gradient}.

\subsection{Mesh Refinement with Spaceborne Pushbroom (Line) Sensors}\label{sec:rpccrefinement}
The de-facto standard to model the object-to-image space mapping for satellite imagery is the RFM. In the following, we derive a mesh refinement scheme like the one above under the RFM (sections \ref{sec:rpc}-\ref{sec:virtcam}). Implementation details for that new model will be given in section \ref{sec:impl_details}. 
\subsubsection{Derivatives of the RFM}\label{sec:rpc}
Since the projection function from object to image space is different from the pinhole model, we first need to adapt the Jacobian $D\Pi(\bfX)$ . Let $B_n=B/s_B+o_B$, $L_n=L/s_L+o_L$ and $H_n=H/s_H+o_H$ be normalised geographic lat/lon/height coordinates of an object point, and let ${\mathbf{P}}(B_n,L_n,H_n)$ be a 20-dimensional vector holding the 3rd-degree polynomial expansion of those normalised coordinates. With the four vectors ${\mathbf{N}}_s, {\mathbf{D}}_s,{\mathbf{N}}_l,{\mathbf{D}}_l$ each holding 20 rational polynomial coefficients (RPC), the image coordinates of the projected point are
\begin{align}
\begin{bmatrix}
           x \\
           y \\
\end{bmatrix}
=
\begin{bmatrix}
           s_s\frac{{\mathbf{N}}^T_{s}{\mathbf{P}}(B_n,L_n,H_n)}{{\mathbf{D}^T}_{s}{\mathbf{P}}(B_n,L_n,H_n)} + o_{l} + o_{cl}\\
           s_l\frac{\mathbf{N}^T_{l}{\mathbf{P}}(B_n,L_n,H_n)}{{\mathbf{D}^T}_{l}{\mathbf{P}}(B_n,L_n,H_n)} + o_{s}+ o_{cs} \\
\end{bmatrix}\;,
\end{align}
with $s_s$, $s_l$ the scales and offsets between pixels and normalised image coordinates. The RFM parameters shipped with an image tend to only be correct up to a small, global bias. That error is often compensated with an additional affine transformation \cite{fraser2004}, or with only a translation \cite{fraser2006}. We follow the latter and apply only a simple shift correction $[o_{cl}, o_{cs}]$. The 2$\times$3 Jacobian matrix $D\Pi(\bfX)$ w.r.t the geographic coordinates reads
\begin{align}
D\Pi(B,L,H)
=
\begin{bmatrix}
           s_s
           \frac
           {\mathbf{N}^T_{s}\mathbf{P}\mathbf{D}^T_{s}}
           {(\mathbf{D}^T_{s}\mathbf{P})^2} 
\begin{bmatrix}             \frac{1}{s_B}\frac{\partial{\mathbf{P}}}{\partial{B_n}} \quad
                            \frac{1}{s_L}\frac{\partial{\mathbf{P}}}{\partial{L_n}} \quad
                            \frac{1}{s_H}\frac{\partial{\mathbf{P}}}{\partial{H_n}}
\end{bmatrix} \\
            s_l 
            \frac
            {\mathbf{N}^T_{l}\mathbf{P}\mathbf{D}^T_{l}}
            {(\mathbf{D}_{l}^T\mathbf{P})^2}
\begin{bmatrix}           \frac{1}{s_B}\frac{\partial{\mathbf{P}}}{\partial{B_n}} \quad
                          \frac{1}{s_L}\frac{\partial{\mathbf{P}}}{\partial{L_n}} \quad
                          \frac{1}{s_H}\frac{\partial{\mathbf{P}}}{\partial{H_n}}
\end{bmatrix}
\end{bmatrix}.
\end{align}

\subsubsection{Quasi-Cartesian Coordinate System}\label{sec:qccos}
The RFM relates Cartesian image coordinates to polar geographic coordinates. For the mesh refinement, it is not only easier, but also numerically advantageous to operate in local Cartesian coordinates. Hence, we transform geographic coordinates $[B, L, H]$ into a "Quasi-Cartesian" local coordinate system. This can be achieved by scaling latitude $B$ and longitude $L$ to the (metric) unit of the height component $H$. Let ${\mathbf{X}}_{geo}=[B_c, L_c, H_c]^T$ be a point located at the center of the area of interest and ${\mathbf{X}}_{utm}=[X_c,Y_c,H_c]^T$ its UTM coordinates. Furthermore, let $[B_{c+1}, L_{c+1}, H_c]^T$ be the corresponding geographic coordinates of $[X_c+1.0,Y_c+1.0, H_c]^T$. Then the transformation to a local, quasi-Cartesian frame can be expressed as 
\begin{equation}
    {\mathbf{X}}_{loc}=f(B,L,H)=\bigg[\frac{B}{B_c-B_{c+1}}, \frac{L}{L_c-L_{c+1}}, H\bigg]^T - {\mathbf{X}}_{geo}.
\end{equation}
This transformation mimics a Cartesian coordinate system only locally, but the approximation is valid for a large enough area (refinement of large areas will in practice always be done for local tiles, in order to parallelise the computation). The quality of the approximation (non-orthogonality and scale anisotropy of the coordinate axes) is shown in Table \ref{table:sys_ortho}. Using the chain rule, the Jacobian of the transformation from image space to quasi-Cartesian coordinates reads
\begin{equation}
    D\Pi(X,Y,H)=
    \begin{bmatrix}    \frac{1}{B_c-B_{c+1}} \quad \frac{1}{L_c-L_{c+1}} \quad 1 \\
                      \frac{1}{B_c-B_{c+1}} \quad \frac{1}{L_c-L_{c+1}} \quad 1
    \end{bmatrix} \odot
    D\Pi(B,L,H),
\end{equation} 
with $ \odot $ denoting element-wise multiplication.
\subsubsection{Independence of Projection Center and Approximation of Depth}\label{sec:independence}
The geometric derivations of (\ref{equ:deltad}) and (\ref{equ:deltaS}) are based on the depth $z$, and on the ray ${\mathbf{d}}$ connecting the projection center and a surface point. The RFM does not model a single projection centre.
Fortunately, (\ref{equ:deltad}) is in fact independent of the absolute length $d=|\mathbf{d}|$. This can be seen by rescaling it to an arbitrary length, e.g., to the unit vector, $\mathbf{d} = d \mathbf{n}_d$. Equation (\ref{equ:deltad}) then becomes
\begin{equation} \label{equ:deltadup}
\frac{\partial\Pi^{-1}_{i,\mathcal{S}+\epsilon\delta\mathcal{S}}(\bfx_i)}{\partial\epsilon}
\Bigg\vert_{\epsilon=0}=\frac{{\mathbf{n}}^T\delta S(\mathbf{X})}{{\mathbf{n}}^T {\mathbf{n}}_d} {\mathbf{n}}_d.
\end{equation}
In contrast, for the variation of image space coordinates $\mathbf{x}_i$ ((\ref{equ:deltaS})) the length of $\mathbf{d}$ does not cancel out easily. Instead, we use the fact that for large focal lengths $d\approx z$. Thus, the denominator is approximately $\frac{1}{z}$. In other words, a stereo model contributes to the gradient inversely proportional to its distance from the surface. We account for this weighting by scaling with the GSD $g$ at the average terrain height. As $\frac{1}{z}\sim g$, this corresponds to a change of variables
\begin{equation} \label{equ:deltaSup}
d{\mathbf{x}}_i=-g {\mathbf{n}}^{T} {\mathbf{n}}_{d,i}d\mathbf{X},
\end{equation}
and the final gradient is calculated as
\begin{equation}\label{equ:satgrad}
\nabla \mathcal{M}_{ij}(S)=-\phi_{\Omega_{\mathcal{S}}}\left[\partial_2M(\mathbf{x}_i)D\bfI_j(\mathbf{x}_j)D\Pi_j(\mathbf{X})g {\mathbf{n}}_{d,i}\right]{\mathbf{n}}.
\end{equation}

\subsubsection{Approximation of Straight Rays With Virtual Cameras} \label{sec:virtcam}
Computing the derivative of the similarity measure $\partial_2M$ requires mapping the image $\bfI_j$ into the view $i$ via the surface, which involves ray casting. Viewing rays modeled by the RFM are in general not straight lines which prevents efficient ray casting. In practice one can assume that the curvature of the rays is low enough to represent them by straight rays in the vicinity of the surface (see Table \ref{table:straight}). Therefore, we define a virtual camera in the following way: We define two virtual planes in object space with constant heights $h$ and $h+ \Delta h$ above ground. Then each image pixel is projected to those two planes using the RFM (and mapped to quasi-Cartesian coordinates), to obtain two virtual points $\mathbf{v}=[x,y,h]$ and $\mathbf{v}'=[x',y',h+\Delta h]$. The set of line segments  
$(\mathbf{v}'-\mathbf{v})$, together with the corresponding pixel intensities, forms a virtual camera that observes the surface along straight rays.

\subsection{Implementation Details}\label{sec:impl_details}
The overall pipeline for surface refinement proceeds as follows:
First, the input mesh is transformed to quasi-Cartesian coordinates. Next, two virtual cameras $\bfI_i^v$, $\bfI_j^v$ 
are set up for  each stereo pair. Now the iterative refinement starts; the image intensities of $\bfI_j^v$ are projected onto the current surface and back into $\bfI_i$. There, we densely compute the similarity (in our implementation ZNCC) between the original and the projected images, as well as its derivative. From $\bfI_i^v$ we read out the ray direction ${\mathbf{n}}_d$. The remaining components needed to compute the gradient (\ref{equ:satgrad}) are obtained as discussed in sections \ref{sec:rpc}-\ref{sec:virtcam}. Per-vertex gradients are obtained by integrating (\ref{equ:satgrad}) over all faces in a one-ring neighborhood. For each vertex of the input mesh, the resulting gradients are summed over all stereo models and scaled with the step size to obtain a field of vertex displacements.
To these photometric gradients, we add displacement vectors corresponding to the thin-plate energy \cite{Kobbelt} to regularise the surface smoothness \cite{vu2012high}. The final displacement field is applied to the mesh vertices to obtain an updated surface, which then serves as input to the next iteration.
Formally, the overall energy is given as
\begin{equation}
E(\mathcal{S})=\alpha \mathcal{M}(\mathcal{S})+\beta \int_{\mathcal{S}}(\kappa_1+\kappa_2)d\mathcal{S},
\end{equation}
where $\kappa1$, $\kappa2$ denote the principal curvatures of the surface. The weight $\alpha$ balances the photometric term and the smoothness. Homogenisation of smoothness and photometric energies is achieved with an additional parameter $\beta=\frac{1}{gsd^2}$ that account for different scales across datasets and mesh resolutions.

In all experiments we run 20 iterations of gradient descent, after which the energy barely decreases any more. More advanced stopping criteria are of course possible. The weight factor $\alpha$ and the step size for gradient decent were derived by grid search. 

 We observed convergence problems for input meshes with
 vertices located too far from the correct surface. We found that convergence can be improved by a hierarchical processing scheme, similar to \cite{li2016efficient}. To that end, we convert the input mesh to a cloud of oriented points and extract a low-resolution mesh via Poisson reconstruction \cite{kazhdan2013screened}. Thereby we choose the minimal voxel resolution (respectively, octree leaf dimension) to $2^l$ GSD.
 The low-resolution mesh is refined using downscaled versions of the original images (factor $2^l$) for 20 iterations. Then the mesh is densified by splitting each triangle face into four smaller ones, and refinement is repeated with image scale $2^{l-1}$, and so forth until the full resolution is reached. Note that the densification factor is the same as the increase in the number of pixels from one pyramid level to the next, hence the (average) number of pixels per triangle remains the same. In our current implementation a triangle covers 2 pixels in average. For additional implementation details we refer the interested reader to our publicly accessible implementation of mesh refinement for pinhole models published in \cite{blaha2017semantically}.
\begin{table}[h] 
\centering
\renewcommand{\arraystretch}{0.7}
\begin{tabular}{| c | c | c | c |}
 \hline 
 scale s & length $|\mathbf{x}'|_2$ [m]& length $|\mathbf{y}'|_2$ [m] & angle $\mathbf{x'}$, $\mathbf{y'} $ [deg] \\ 
 \hline
 100 & 100.002 & 100.002 & 90.000 \\  
 
 200 & 200.003 & 200.003 & 90.001 \\

 500 & 500.009 & 500.008 & 90.001 \\

 1000 & 1000.018 & 1000.016 & 90.003 \\ 
 
 2000 & 2000.038 & 2000.031 & 90.005 \\

 5000 & 5000.110 & 5000.067 & 90.014 \\
\hline
\end{tabular}   
    \caption{Approximation error of the quasi-Cartesian local coordinate system. To check orthogonality we define two orthogonal unit vectors $\mathbf{x}=\mathbf{p}_1-\mathbf{p}_0$ and $\mathbf{y}=\mathbf{p}_2-\mathbf{p}_0$ in  UTM coordinates.
    $\mathbf{p}_0$ is located in the center of the test area described in section \ref{sec:sites}. The points $\mathbf{p}_{0,1,2}$ are located in a horizontal plane at the average terrain height. We transform scaled versions $s\mathbf{x}$ ad $s\mathbf{y}$ to geographic coordinates and then to the quasi-Cartesian frame to obtain $\mathbf{x}'$ and $\mathbf{y}'$. For vectors of length 2000m the scale difference in  $x$- and $y$-direction is $<$7mm, the deviation of $x$- and $z$-axis is 38mm, or 0.13 times the GSD of the best available civilian satellite imagery, which is well below the absolute precision of the RPC projection. The angle between $\mathbf{x}'$ and $\mathbf{y}'$ is $<$0.006$^\circ$
}\label{table:sys_ortho}
\end{table}

\begin{table}[h] 
\centering
\renewcommand{\arraystretch}{0.7}
\begin{tabular}{| c | c | c | c | c |}
 \hline
 h[m] &  1  & 100 & 500 & 1000 \\ %
 \hline
 $\phi$ [$^{\circ}$] &  30.131 & 30.130 & 30.128 & 30.126 \\ \hline %

\end{tabular}   
    \caption{Approximation of off-nadir ray angles $\phi$. The table displays the direction of a ray close to the surface, approximated with our virtual camera construction. By varying $h$ it can be seen that the rays are indeed slightly curved. However, the difference between $h=0m$ and $h=1000m$ is $\approx$ 0.005 $^{\circ}$. In other words, the error of the virtual camera approximation is negligible for any reasonable camera height $h$.
    }
    \label{table:straight}
\end{table}

\section{Results}

\subsection{Test Sites}\label{sec:sites}

\begin{figure}[t]
\centering
\includegraphics[height=4.6cm]{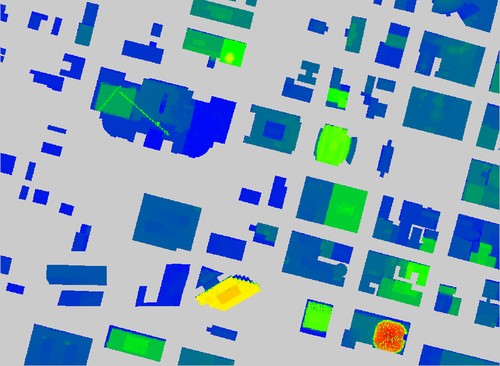}
\hspace{0.5cm}
\includegraphics[height=4.6cm]{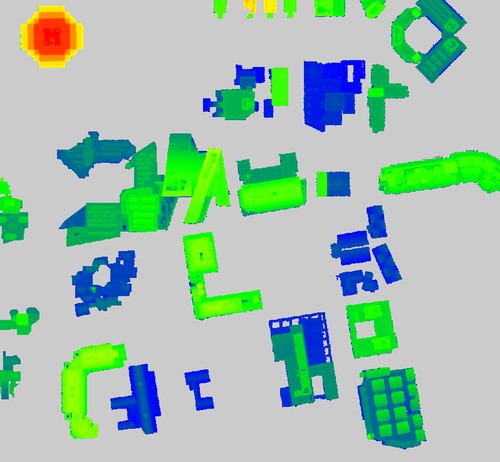}
\caption{Ground truth LiDAR DSM data of evaluated building areas of JAX test site (left) and UCSD test site (right).}
\label{fig:evaluation_area}
\end{figure}

\begin{figure}[t]
\centering
\begin{subfigure}{0.3\textwidth}
\includegraphics[width=\linewidth]{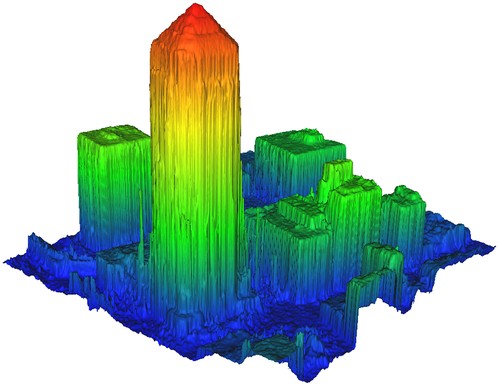}
\caption{} \label{fig:pipeline_a}
\end{subfigure}
\begin{subfigure}{0.3\textwidth}
\includegraphics[width=\linewidth]{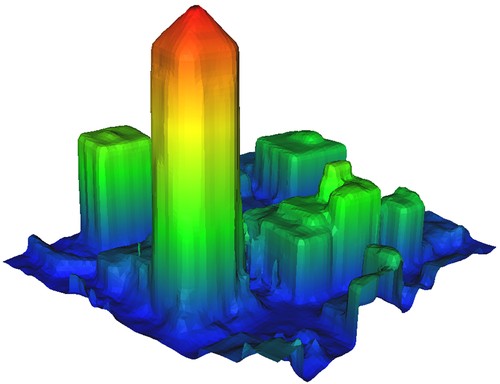}
\caption{} \label{fig:pipeline_b}
\end{subfigure}
\begin{subfigure}{0.3\textwidth}
\includegraphics[width=\linewidth]{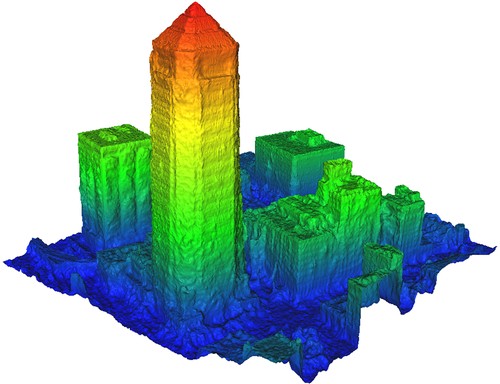}
\caption{} \label{fig:pipeline_c}
\end{subfigure}
\caption{Example of (a) the DSM mesh (b) the input Poisson mesh (c) the refined mesh for region of interest (JAX).}
\label{fig:pipeline}
\end{figure}

We test the proposed algorithm on the publicly available benchmark  \cite{brown2018large}, which provides multi-view collections of 16bit panchromatic WorldView-3 images with 0.3m GSD (at nadir,
the actual GSD in off-nadir views can be up to a factor of $\approx$1.5 lower).
Two different test sites were evaluated: Downtown Jacksonville (JAX), FL, USA and University of California San Diego (UCSD), CA, USA. 
The JAX test site features 26 images of an urban area of ca. 750m$\times$750m, collected between October 2014 and February 2016. The UCSD test site consists of 35 images covering an area of ca.\ 600m$\times$600m, collected between October 2015 and August 2017.
A 2.5D LiDAR DSM with 0.5m GSD is provided and serves as ground truth for both test sites, see Figure \ref{fig:evaluation_area}. The lack of full-3D ground truth makes it impossible to quantitatively evaluate 3D elements, such as indentations on facades, in our reconstructed mesh, see Figure \ref{fig:pipeline}(c).
Nevertheless, since the gist of our approach is its ability to reconstruct true 3D geometry, we first present a qualitative evaluation of such 3D areas in the following section, while the quantitative evaluation based on 2.5D elevation maps is discussed in section \ref{sec:quant}.  

\subsection{Generation of Initial Meshes, Pre-Processing and Parameters}\label{sec:preproc}
The proposed mesh refinement needs an initial surface mesh to start from. We generate that mesh by running conventional dense stereo and meshing the resulting 2.5D elevation model by Poisson reconstruction (Section \ref{sec:impl_details}). The MVS system requires sub-pixel accurate relative alignment for good performance, which is not reached by the image provider's RPCs. We use RFM parameters that have been refined with the method of \cite{avibench}, which implements bias-correction via feature matching and subsequent bundle block adjustment. 

The image collections feature very high redundancy, almost all images of a site have significant overlap. It would not be meaningful to use possible pairings for surface reconstruction: not only would it be computationally expensive, it can also be harmful to include images with baselines too short for accurate triangulation or too long for robust matching. According to our experimental experience and related research \cite{dAngelo2012, Facciolo2017, Qin2017}, we manually remove images with overly poor illumination conditions and select 80 suitable image pairs for JAX and 86 for UCSD, with intersection angles of the viewing directions between $5^\circ$ and $13^\circ$. The parameter $\alpha$ which controls the contribution of the unary term was set to $3.5\cdot10^4$ for UCSD and $4\cdot10^4$ for JAX respectively. The parameter $\beta$, which steers the smoothness was set to 0.05 for both datasets. The average size of projected triangles is 2 pixels. Meshes were refined using the coarse-to-fine scheme starting at $1/8$ resolution and stopping at the full resolution images. With the current, unoptimised implementation the runtime for the full-resolution refinement is 65mins for JAX and 39mins for the samller UCSD. The most time consuming part is ray casting, which consumes $\approx$70\% of the computation time. We note that the ray casting is suitable for GPU implementation, as is the computation of $\partial M(\mathbf{x}_i)$, which furthermore could be reused and updated only periodically after several iterations. Together with stricter mechanisms for stereo-pair selection and masking of regions outside the stereo overlap, a $>$10$\times$ speed-up is almost certainly possible.

\subsection{Qualitative Evaluation - Reconstruction of 3D Structure}\label{sec:qual}

\begin{figure}[htbp]
\centering
\begin{subfigure}{0.28\textwidth}
\includegraphics[width=\linewidth]{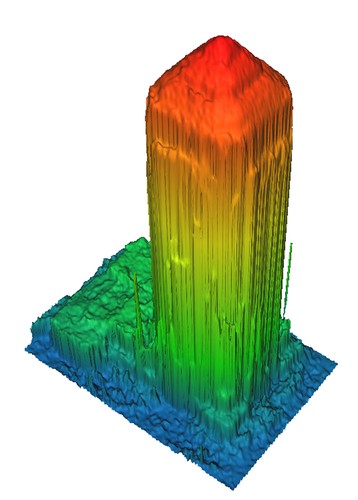}
\caption{} \label{fig:roi3_a}
\end{subfigure}
\begin{subfigure}{0.32\textwidth}
\includegraphics[width=\linewidth]{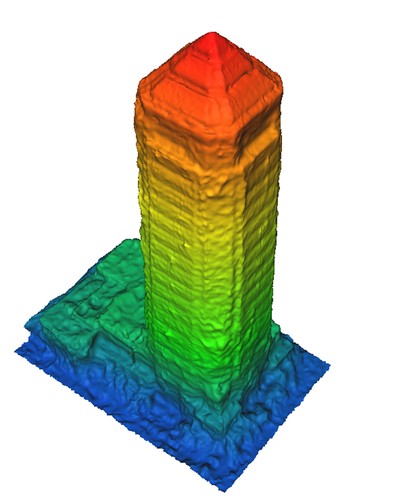}
\caption{} \label{fig:roi3_b}
\end{subfigure}
\begin{subfigure}{0.32\textwidth}
\includegraphics[width=\linewidth]{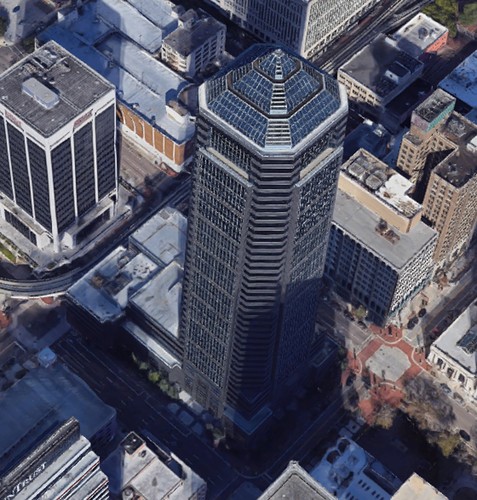}
\caption{} \label{fig:roi3_c}
\end{subfigure}
\begin{subfigure}{0.32\textwidth}
\includegraphics[width=\linewidth]{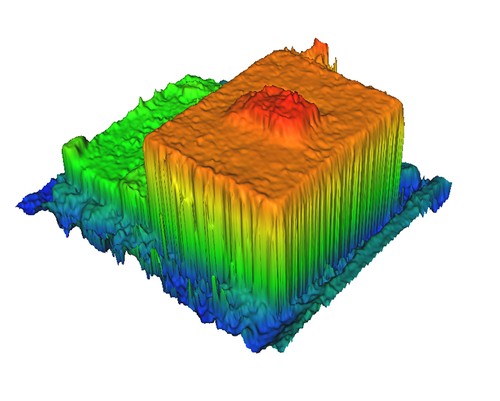}
\caption{} \label{fig:roi1_a}
\end{subfigure}
\begin{subfigure}{0.32\textwidth}
\includegraphics[width=\linewidth]{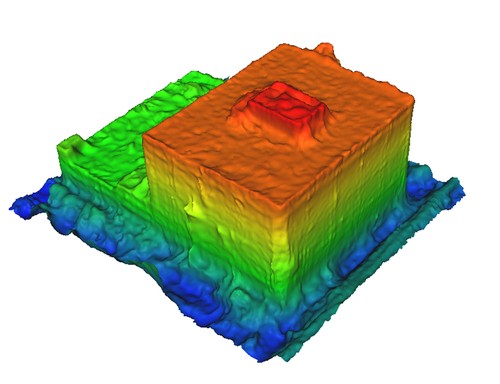}
\caption{} \label{fig:roi1_b}
\end{subfigure}
\begin{subfigure}{0.32\textwidth}
\includegraphics[width=\linewidth]{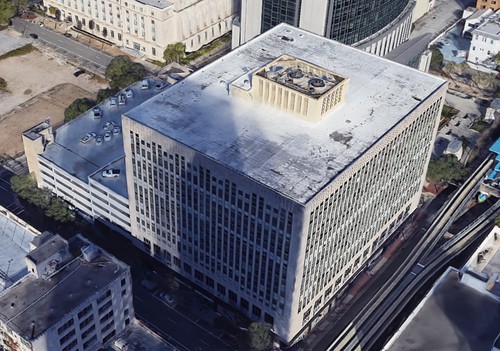}
\caption{} \label{fig:roi1_c}
\end{subfigure}
\begin{subfigure}{0.32\textwidth}
\includegraphics[width=\linewidth]{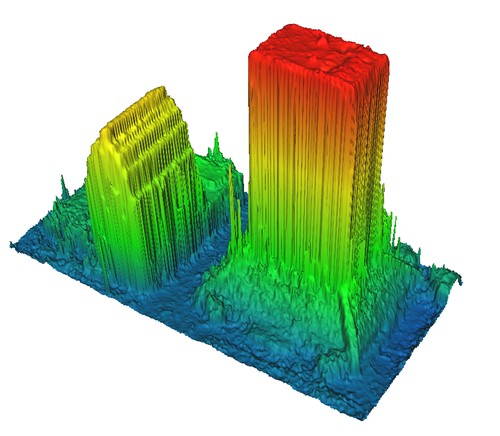}
\caption{} \label{fig:roi2_a}
\end{subfigure}
\begin{subfigure}{0.32\textwidth}
\includegraphics[width=\linewidth]{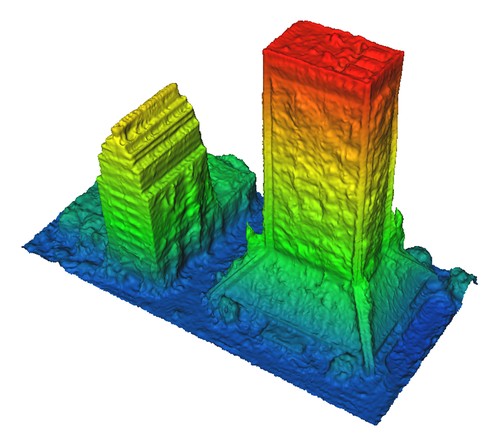}
\caption{} \label{fig:roi2_b}
\end{subfigure}
\begin{subfigure}{0.32\textwidth}
\includegraphics[width=\linewidth]{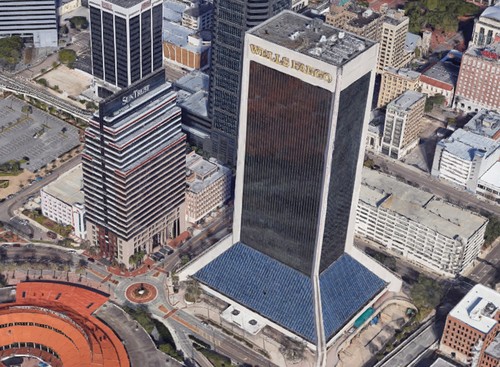}
\caption{} \label{fig:roi2_c}
\end{subfigure}
\begin{subfigure}{0.28\textwidth}
\includegraphics[width=\linewidth]{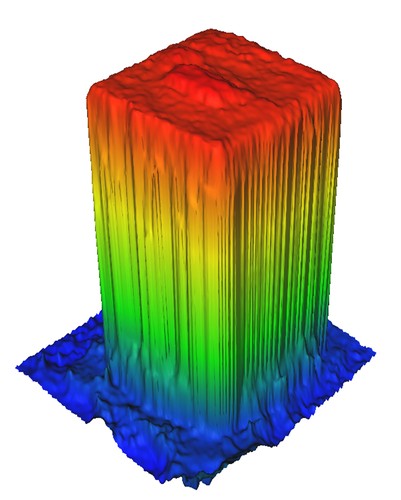}
\caption{} \label{fig:roi4_a}
\end{subfigure}
\begin{subfigure}{0.32\textwidth}
\includegraphics[width=\linewidth]{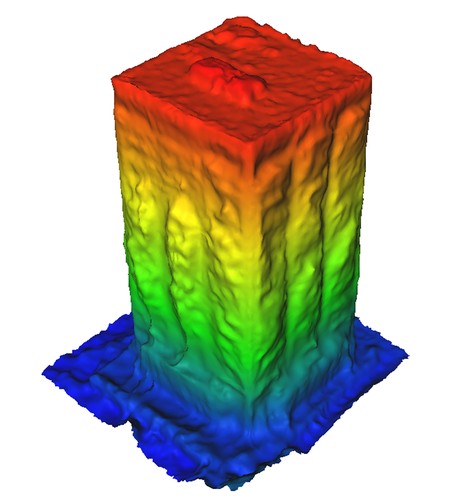}
\caption{} \label{fig:roi4_b}
\end{subfigure}
\begin{subfigure}{0.32\textwidth}
\includegraphics[width=\linewidth]{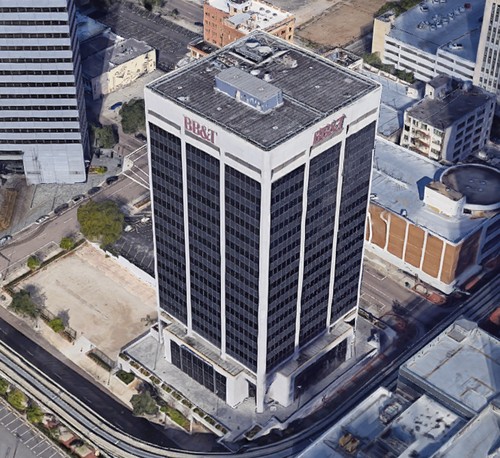}
\caption{} \label{fig:roi4_c}
\end{subfigure}
\caption{Visualization of building surfaces generated by tSGM (left column), the refined versions (middle column) and Google Maps snapshots (right column).}
\label{figure:3d_eval}
\end{figure}

In this section we qualitatively assess the reconstructions obtained with the proposed 3D mesh refinement algorithm. In particular, we illustrate its ability to recover 3D shape details that are not representable in a 2.5D height field, and its superior treatment of sharp discontinuities on man-made structures.
Figure \ref{figure:3d_eval} shows 2.5D tSGM models (\ref{fig:roi3_a}, \ref{fig:roi1_a}, \ref{fig:roi2_a}, \ref{fig:roi4_a}), the refined 3D models (\ref{fig:roi3_b}, \ref{fig:roi1_b}, \ref{fig:roi2_b}, \ref{fig:roi4_b}) and Google Maps snapshots (\ref{fig:roi3_c}, \ref{fig:roi1_c}, \ref{fig:roi2_c}, \ref{fig:roi4_c}) of four example regions of interest (ROIs). Facades of the building displayed in figure \ref{fig:roi1_b} are clearly smoother than in the 2.5D version displayed in \ref{fig:roi1_a}. For 2.5D models, the facade geometry is defined by roof and ground elevations. Errors in such elevation estimates are propagated over the whole facade. 3D refinement in facade regions is supported by image similarity in oblique views, leading to smoother surfaces without raster artifacts, and without destroying high-frequency crease edges. Thus, the refined models have visibly crisper crease edges.
On the building in the first row, there are improvements of the roof geometry (little steps). Such shapes are inherently difficult for  2.5D approaches, where these elements are represented only by very few pixels of the elevation map. Moreover, the facades feature indentations and vertical edges that are, by construction, not representable in a 2.5D heightfield.
Similar observations hold for the buildings in the second and third rows: Roof substructures are crisper and more geometric detail is extracted on facades. However, the facades are  overly bumpy in places, presumably due to repetitive structures, specular materials and insufficient evidence in the image set due to the uneven distribution of viewing directions.
For the challenging case in the fourth row vertical roof structures are only partly reconstructed, apparently due to weak data support. Also the large, dark glass surfaces again impair the facade reconstruction.

\begin{figure}[t]
\centering
\begin{subfigure}{0.32\textwidth}
\includegraphics[width=\linewidth]{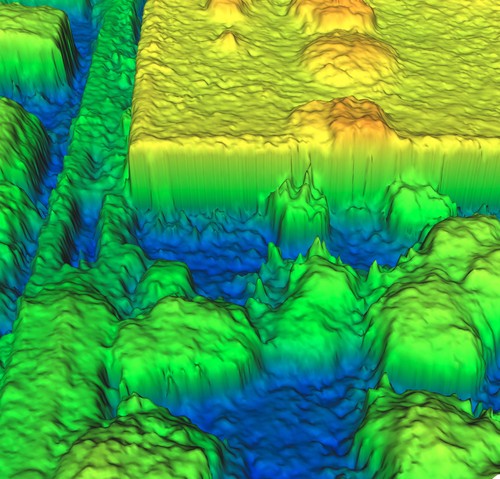}
\caption{} \label{fig:roi5_a}
\end{subfigure}
\begin{subfigure}{0.32\textwidth}
\includegraphics[width=\linewidth]{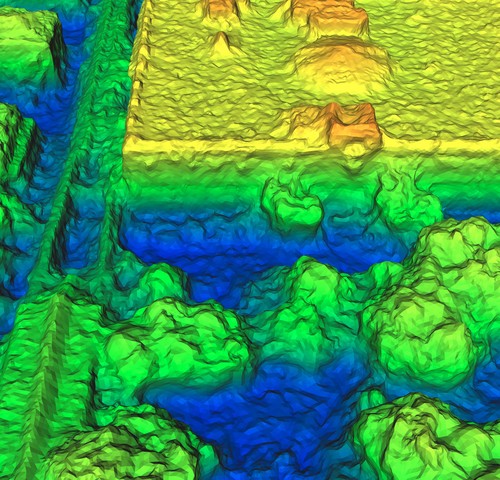}
\caption{} \label{fig:roi5_b}
\end{subfigure}
\begin{subfigure}{0.32\textwidth}
\includegraphics[width=\linewidth]{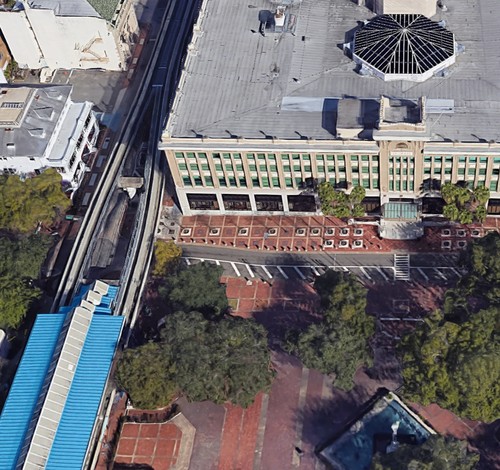}
\caption{} \label{fig:roi5_c}
\end{subfigure}
\begin{subfigure}{0.32\textwidth}
\includegraphics[width=\linewidth]{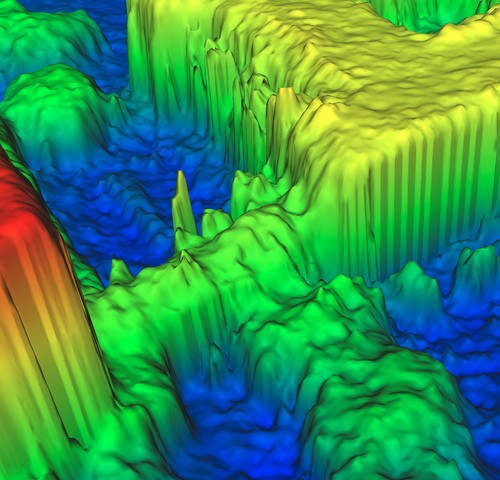}
\caption{} \label{fig:roi6_a}
\end{subfigure}
\begin{subfigure}{0.32\textwidth}
\includegraphics[width=\linewidth]{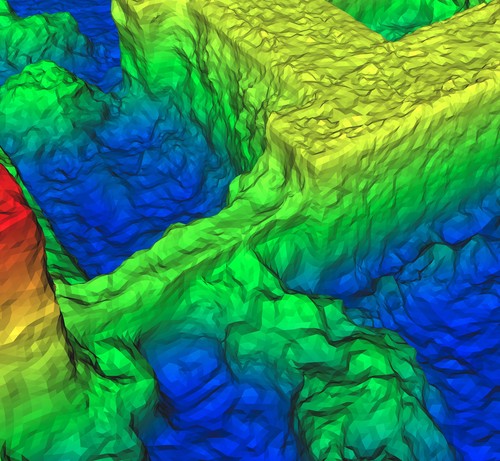}
\caption{} \label{fig:roi6_b}
\end{subfigure}
\begin{subfigure}{0.32\textwidth}
\includegraphics[width=\linewidth]{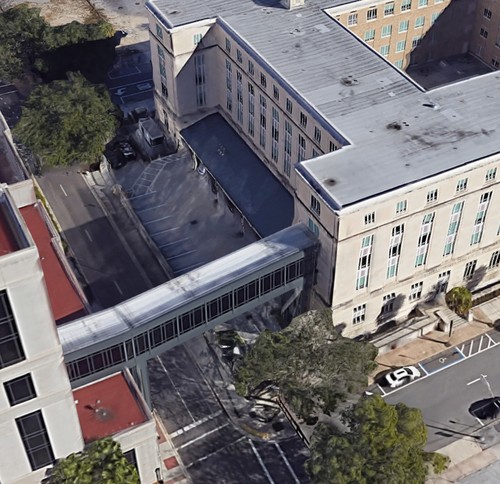}
\caption{} \label{fig:roi6_c}
\end{subfigure}
\caption{Visualization of general surfaces generated by tSGM (left column), the refined versions (middle column) and Google Maps snapshots (right column).}
\label{figure:3d_eval2}
\end{figure}

Figures \ref{fig:roi5_a}-\ref{fig:roi6_c} depict surfaces of vegetation, bridges, thin roof structures and railway tracks. Compared to tSGM, refined surfaces offer more detail, feature  less outliers and appear less noisy. The same holds true for reconstructed vegetation. The street under the bridge (Figure \ref{fig:roi6_b}) is not reconstructed correctly -- while one can see the attempt to "carve out" the empty space under the bridge, the refinement is limited by the faulty topology of the initial surface, which cannot be changed by the algorithm.

\subsection{Quantitative Evaluation of Multi-View Refinement}
\label{sec:quant}%
To test the sensitivity of our method w.r.t.\ the initialisation, we generate DEMs with three different state-of-the-art satellite MVS systems: (a) \cite{Gong2019} using a hierarchical version of SGM \cite{rothermel2012sure}, (b) the S2P pipeline \cite{Facciolo2017} based on MGM \cite{facciolo2015mgm}, and (c) NASA's Stereo Pipeline (ASP) \cite{nasa_asp} based on Semi-Global-Block-Matching \cite{opencv_library}.
The three DEMs also serve as baselines to compare with. For a fair comparison we use the refined RFM parameters in all MVS systems and in the actual mesh refinement.

The LiDAR ground truth is provided in the form of a 2.5D gridded DEM. Consequently, the refined 3D meshes must be converted back to 2.5D elevation maps. To accomplish this process, we align the mesh to the LiDAR DEM \cite{Bosch2016}, cast a vertical ray through the centre of each grid cell, and extract the highest intersection point with the reconstructed mesh.

\subsubsection{ Evaluation of Refined Surfaces in Benchmark Regions}\label{sec:numevalall}
Table \ref{table:evaluate_all} displays error statistics for the MVS results and the corresponding refinement results, for both test sites. The comparisons to ground truth were carried out with the test suite provided by \cite{Bosch2016}, where we add additional, robust error metrics: namely,  a truncated root mean square error RMSE, computed from only those residuals that are $<$3m (ca. 10GSD), and the corresponding completeness (percentage of residuals below that threshold). Furthermore, we list the normalised median absolute residuals and the 68\% percentile of absolute residuals, denoted by perc-68. This value is calculated such, that 68\% of the residuals are $\leq\,$perc-68, corresponding to $\pm\, 1\!\cdot\!\sigma$ of a Gaussian error distribution.
Note that the framework evaluates only on selected building roofs, see Figure~\ref{fig:evaluation_area}, to rule out error sources like seasonal changes, extreme height discontinuities, temporal changes and moving objects. \cite{brown2018large} provide a building mask for the Jacksonville test site. Since for the UCSD site the building mask is not publicly available, we manually created one. As shown in table \ref{table:evaluate_all}, our mesh refinement significantly improves the 2.5D DEMs generated with both the SGM and ASP methods.
There are no significant quantitative differences to S2P (for JAX the refined version is insignificantly better, for UCSD insignificantly worse). We note that the error metrics do not fully reflect the visual quality of the reconstructions. The mesh refinement does bring out additional 3D structure and suppress noise (see Sec.~\ref{sec:eval_detail}), but it appears that in terms of average 2.5D roof accuracy the S2P method is already close to the achievable limit (normalised median absolute deviation $<$1.5$\,$GSD), so that no further improvement is possible.
In general, the accuracy of the DEMs, at least on roof surfaces, is in the sub-metre range, which we find very encouraging. The majority of the residuals are $<$1.5$\,$GSD.
Note that the mesh refinement exhibits little sensitivity to the initialisation, its results are quantitatively very similar in all cases, independent of which stereo method was used to generate the input surface.
\begin{table}
\begin{center}
\centering
\renewcommand{\arraystretch}{0.7}
\begin{tabular}{| c | c | c | c | c | c | c | c |}
	\hline
		& &\multicolumn{1}{c|}{tSGM} & \multicolumn{1}{c|}{tSGM ref} & \multicolumn{1}{c|}{S2P} & \multicolumn{1}{c|}{SP2 ref} & \multicolumn{1}{c|}{ASP} & \multicolumn{1}{c|}{ASP ref}\\
 \hline \hline
 \parbox[t]{2mm}{\multirow{4}{*}{\rotatebox[origin=c]{90}{\textbf{JAX}}}}
   & Compl. [\%] &  88.46 & \textbf{88.77} & 88.32 & \textbf{88.39} & 84.22 & \textbf{87.14} \\ 
   & RMSE [m] & 0.86 & \textbf{0.80} & 0.79 & \textbf{0.78} & 1.0 & \textbf{0.83}  \\ 
   & NMAD [m]  & 0.51 & \textbf{0.39} & 0.42 & \textbf{0.40} & 0.74 & \textbf{0.46} \\ 
   & perc-68 [m] & 0.88 & \textbf{0.73} & 0.72 & \textbf{0.71} & 1.25 & \textbf{0.80}  \\ \hline 
	
 \parbox[t]{2mm}{\multirow{4}{*}{\rotatebox[origin=c]{90}{\textbf{UCSD}}}}
&Compl. [\%] & 95.69 & \textbf{96.09}  & \textbf{96.78} & {96.12} & 93.23 &  \textbf{96.44} \\  
 &  RMSE [m] & 0.97 & \textbf{0.86} & \textbf{0.82} & 0.85 & 1.16 & \textbf{0.86} \\ 
  & NMAD [m]  & 0.58 & \textbf{0.47} & \textbf{0.45} & 0.46 & 0.79 & \textbf{0.47} \\ 
  & perc-68 [m] & 0.95 & \textbf{0.77}  & \textbf{0.74} & 0.75 &  1.28 & \textbf{0.77}\\ 
  \hline
    \end{tabular}
    \captionof{table}{Evaluation results of the JAX and UCSD test site for three MVS methods (\cite{Gong2019,Facciolo2017,nasa_asp}) and corresponding refined surfaces (marked by 'ref').}
\label{table:evaluate_all}
\end{center}
\vspace{-7mm}
\end{table}

\subsubsection{In-depth Evaluation of Refined Surfaces on Individual Buildings} \label{sec:eval_detail}
To characterise the reconstruction in more detail and gain further insights into its behaviour, we examine four building roofs for each test site in detail. As explained in section \ref{sec:numevalall} we exclude building edges, since aliasing at large height jumps makes a meaningful evaluation impossible.
\begin{table}
\begin{center}
\centering
\renewcommand{\arraystretch}{0.7}
\begin{tabular}{ | c | c | c | c | c | c | c | c |}
	\hline 
	& &\multicolumn{1}{c|}{tSGM} & \multicolumn{1}{c|}{tSGM ref} & \multicolumn{1}{c|}{S2P} & \multicolumn{1}{c|}{SP2 ref} & \multicolumn{1}{c|}{ASP} & \multicolumn{1}{c|}{ASP ref}\\
 \hline \hline
 \parbox[t]{1mm}{\multirow{4}{*}{\rotatebox[origin=c]{90}{\textbf{ROI 1}}}}  & Compl. [\%] & 92.69 & \textbf{93.35} & \textbf{93.82} & 92.83 &  89.76 & \textbf{90.57} \\ 
  & RMSE [m] & 0.80 & \textbf{0.79} & \textbf{0.67} & 0.77 & 0.85 & \textbf{0.76} \\ 
  & NMAD [m]  & 0.31& \textbf{0.24} &\textbf{0.22}  & 0.23 & 0.46& \textbf{0.25} \\ 
  & perc-68 [m] & 0.70 & \textbf{0.64}  & \textbf{0.51} & 0.56& 0.85 & \textbf{0.61} \\ \hline
	
\parbox[t]{1mm}{\multirow{4}{*}{\rotatebox[origin=c]{90}{\textbf{ROI 2}}}}
	&Compl. [\%] & 94.69 & \textbf{94.71} & \textbf{95.06} & 94.54 & 89.71 & \textbf{93.49} \\  
   &RMSE [m] & 0.77 & \textbf{0.75} & \textbf{0.69} & 0.74  & 1.07 & \textbf{0.75} \\ 
& NMAD [m]  & 0.37 & \textbf{0.28} & \textbf{0.27} &0.30  & 0.67& \textbf{0.30} \\ 
  & perc-68 [m] &0.71  & \textbf{0.64}  & \textbf{0.60} & 0.62 & 1.22 & \textbf{0.65} \\ \hline 

\parbox[t]{1mm}{\multirow{4}{*}{\rotatebox[origin=c]{90}{\textbf{ROI 3}}}}
&	Compl. [\%] & 79.31 & \textbf{80.54} & \textbf{83.79} & 80.86 & 63.26 & \textbf{80.37} \\  
 &  RMSE [m] & 1.26 & \textbf{1.18} & 1.15 & \textbf{1.13} & 1.56 & \textbf{1.16} \\ 
  &  NMAD [m]  & 1.18&  \textbf{1.00}& 0.88 & \textbf{0.87} &2.15 & \textbf{0.92} \\ 
  & quant-68 [m] & 1.93 & \textbf{1.75} & \textbf{1.52}  &1.68 &3.32 & \textbf{1.78}\\ \hline 
	
\parbox[t]{1mm}{\multirow{4}{*}{\rotatebox[origin=c]{90}{\textbf{ROI 4}}}}
&	Compl. [\%] & 93.68 & \textbf{95.52} & 94.63 & \textbf{94.86} & 88.81 & \textbf{92.64}\\  
 &  RMSE [m] & 0.89 & \textbf{0.76}& 0.82 &\textbf{0.74}  & 1.05 & \textbf{0.75} \\ 
 &  NMAD [m]  & 0.46 & \textbf{0.36} & 0.36 & \textbf{0.35} &0.67 & \textbf{0.36}\\ 
 &  perc-68 [m] &0.84  &\textbf{0.66}  &0.68  &\textbf{0.63} & 1.20& \textbf{0.68}\\ \hline
    \end{tabular}
    \captionof{table}{Evaluation results of the JAX test site for three MVS methods (\cite{Gong2019,Facciolo2017,nasa_asp}) and corresponding refined surfaces (marked by 'ref').}
\label{table:evaluate_JAX_num}
\end{center}
\vspace{-7mm}
\end{table}
\begin{table}
\begin{center}
\centering
\renewcommand{\arraystretch}{0.7}
\begin{tabular}{ | c | c | c | c | c | c | c | c |}
	\hline
	& &\multicolumn{1}{c|}{tSGM} & \multicolumn{1}{c|}{tSGM ref} & \multicolumn{1}{c|}{S2P} & \multicolumn{1}{c|}{SP2 ref} & \multicolumn{1}{c|}{ASP} & \multicolumn{1}{c|}{ASP ref}\\
 \hline \hline
 \parbox[t]{1mm}{\multirow{4}{*}{\rotatebox[origin=c]{90}{\textbf{ROI 1}}}}
   & Compl. [\%] & \textbf{99.70} & 99.20 & 98.50  & \textbf{99.0} & 94.40 & \textbf{99.04}\\  
  & RMSE [m] & 0.97 &\textbf{0.87}  & \textbf{0.86} & 0.88 & 1.21 & \textbf{0.86} \\ 
  & NMAD [m] & 0.61 & \textbf{0.50}  & \textbf{0.48}  & 0.52 & 0.90& \textbf{0.51} \\ 
  & perc-68 [m] & 0.90 & \textbf{0.74} & \textbf{0.74} & 0.75&1.34 & \textbf{0.73} \\ \hline 
	
	 \parbox[t]{1mm}{\multirow{4}{*}{\rotatebox[origin=c]{90}{\textbf{ROI 2}}}}
  & Compl. [\%] & 92.46  & \textbf{92.22} & \textbf{94.17}  & 92.33 & 86.15 & \textbf{91.80}\\  
  & RMSE [m] & 1.12 & \textbf{0.97} & \textbf{0.96}  & 0.97  &1.24 & \textbf{0.98} \\ 
  & NMAD [m]  & 0.68  & \textbf{0.52} & 0.48 & \textbf{0.52} & 0.90 & \textbf{0.54} \\
  & perc-68 [m] & 1.22 & \textbf{0.93} & 0.92 & 0.92  & 1.67 & \textbf{0.96} \\ \hline 
	 \parbox[t]{1mm}{\multirow{4}{*}{\rotatebox[origin=c]{90}{\textbf{ROI 3}}}}
  & Compl. [\%] & 96.00 & \textbf{98.67} & 97.66 & \textbf{98.33} & 85.21 & \textbf{98.31} \\ 
  & RMSE [m] & 1.10 & \textbf{1.07} & \textbf{1.09} & 1.10 & 1.65 & \textbf{1.12}\\ 
    & NMAD [m]  & 0.69 & \textbf{0.59}  & 0.67 & \textbf{0.61} & 1.60 & \textbf{0.69} \\ 
   & perc-68 [m] & 1.09 & \textbf{1.06} & 1.10 & 1.10 & 2.27 & \textbf{1.17} \\ \hline 
   	
	 \parbox[t]{1mm}{\multirow{4}{*}{\rotatebox[origin=c]{90}{\textbf{ROI 4}}}}
 &  Compl. [\%] & 99.14 & \textbf{99.44} & 99.53 & \textbf{99.52} & 96.44  & \textbf{99.31} \\  
 &  RMSE [m] & 0.95 & 0.95& \textbf{0.88} & 0.94 &1.35 & \textbf{0.96} \\ 
  &  NMAD [m] & 0.57& \textbf{0.46} & 0.46 & \textbf{0.45} & 0.70 & \textbf{0.43} \\ 
  & perc-68 [m] & 0.92 & \textbf{0.85} & \textbf{0.83} & 0.84 & 1.48 & \textbf{0.87} \\ \hline

 \end{tabular}
    \captionof{table}{Evaluation results of the UCSD test site for three MVS methods (\cite{Gong2019,Facciolo2017,nasa_asp}) and corresponding refined surfaces (marked by 'ref').} 
\label{table:evaluate_UCSD_num}
\end{center}
\vspace{-6mm}
\end{table}

\begin{table}
\begin{center}
\begin{tabular}{ @{}| c @{\hspace{1mm}}|@{\hspace{1mm}} c @{\hspace{1mm}}|@{\hspace{1mm}} c @{\hspace{1mm}}|@{\hspace{1mm}} c @{\hspace{1mm}}|@{\hspace{1mm}} c @{\hspace{1mm}}|@{\hspace{1mm}} c @{\hspace{1mm}}|@{\hspace{1mm}} c @{\hspace{1mm}}|@{\hspace{1mm}} c @{\hspace{1mm}}|@{}}	\hline
	&LiDAR & tSGM & tSGM ref & S2P & SP2 ref & ASP & ASP ref.\\ \hline
      \rotatebox[origin=lc]{90}{ROI1}
     &\includegraphics[width=1.75cm,angle=90,origin=c]{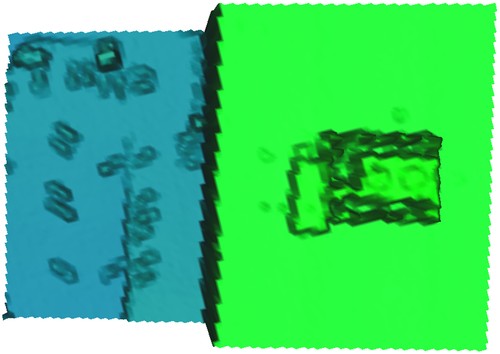}
    & \includegraphics[width=1.75cm,angle=90,origin=c]{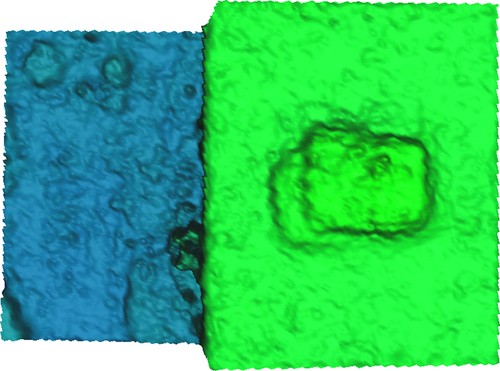}
    & \includegraphics[width=1.75cm,angle=90,origin=c]{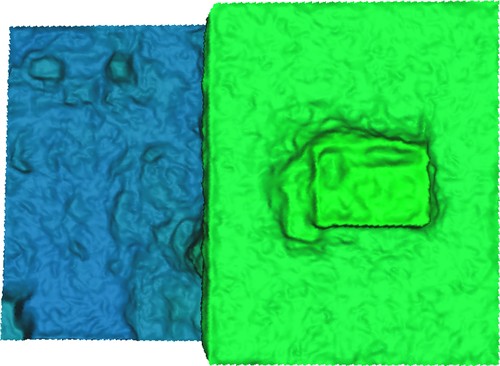}
    & \includegraphics[width=1.75cm,angle=90,origin=c]{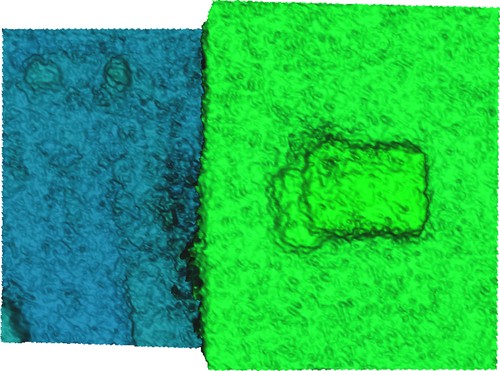}
    & \includegraphics[width=1.75cm,angle=90,origin=c]{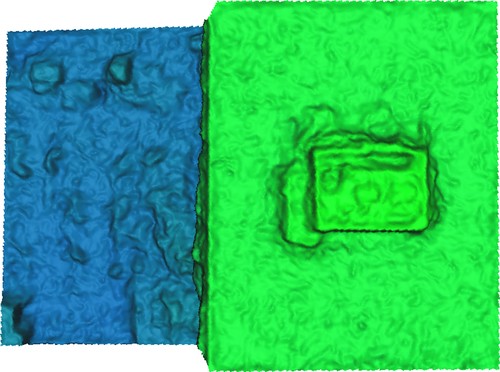}
    & \includegraphics[width=1.75cm,angle=90,origin=c]{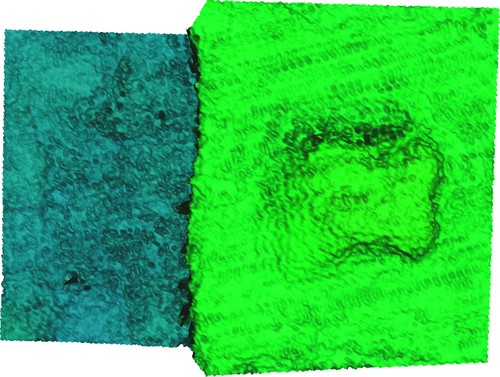}
    & \includegraphics[width=1.75cm,angle=90,origin=c]{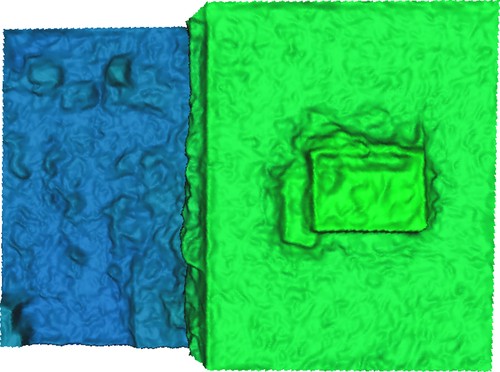}
 \\ \hline
 \rotatebox[origin=lc]{90}{ROI2}
    & \includegraphics[width=1.65cm,angle=90,origin=c]{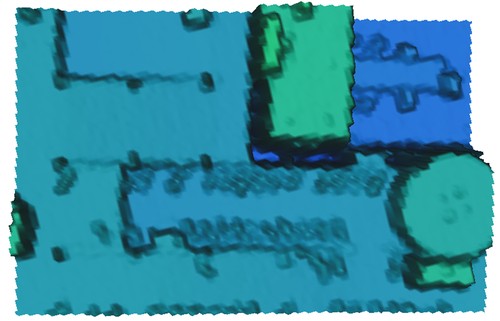}
    & \includegraphics[width=1.65cm,angle=90,origin=c]{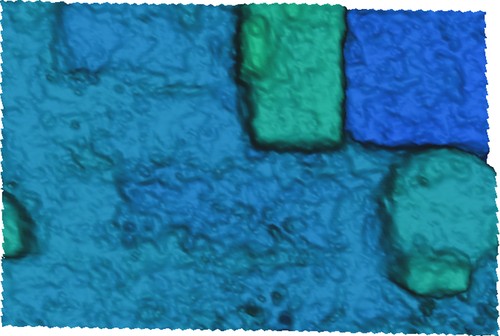}
    & \includegraphics[width=1.65cm,angle=90,origin=c]{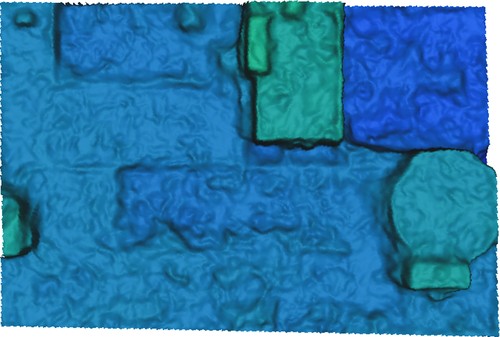}
    & \includegraphics[width=1.65cm,angle=90,origin=c]{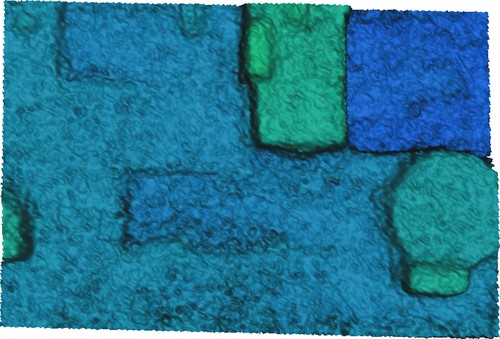}
    & \includegraphics[width=1.65cm,angle=90,origin=c]{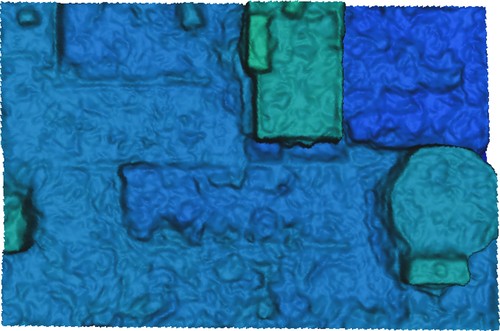}
    & \includegraphics[width=1.65cm,angle=90,origin=c]{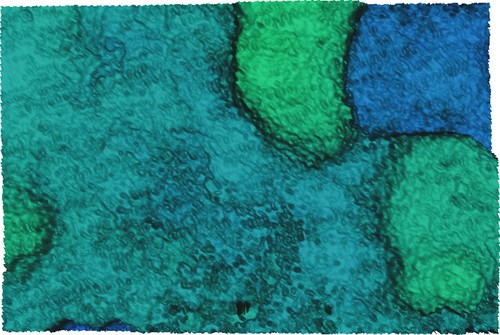}
    & \includegraphics[width=1.65cm,angle=90,origin=c]{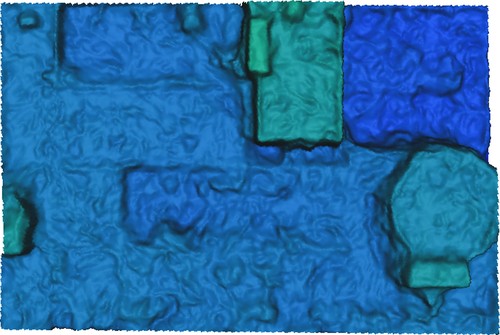}
 \\ \hline
  \rotatebox[origin=lc]{90}{ROI3}
    & \includegraphics[width=1.2cm,origin=c]{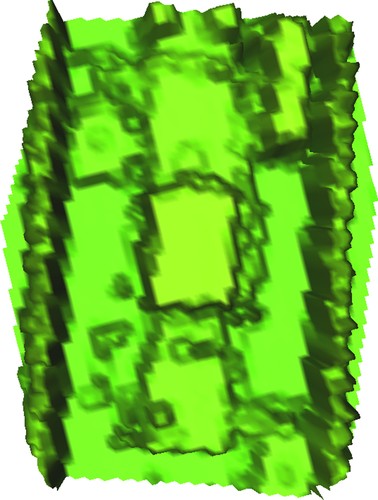}
    &  \includegraphics[width=1.2cm,origin=c]{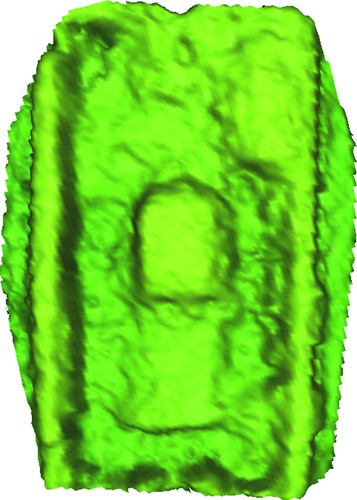}
    & \includegraphics[width=1.2cm,origin=c]{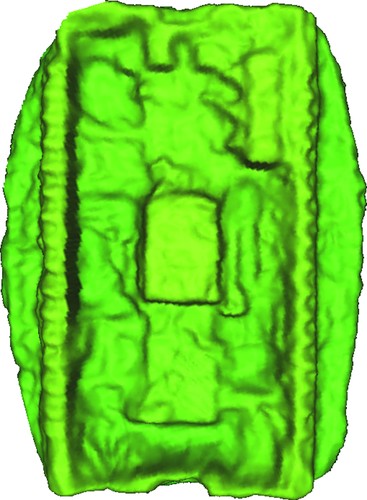}
    & \includegraphics[width=1.2cm,origin=c]{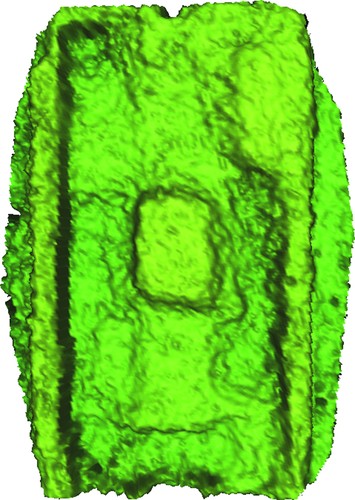}
    & \includegraphics[width=1.2cm,origin=c]{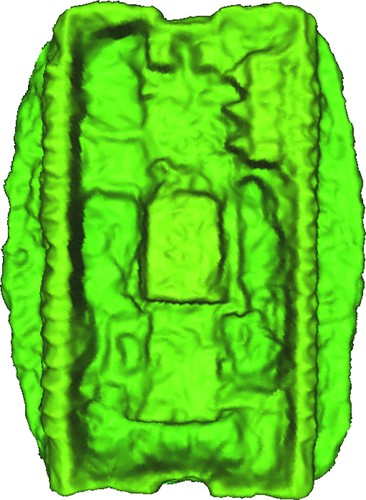}
    & \includegraphics[width=1.2cm,origin=c]{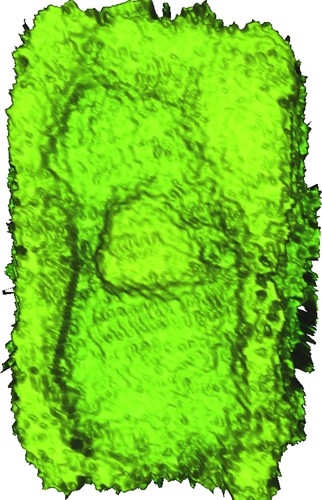}
    & \includegraphics[width=1.2cm,origin=c]{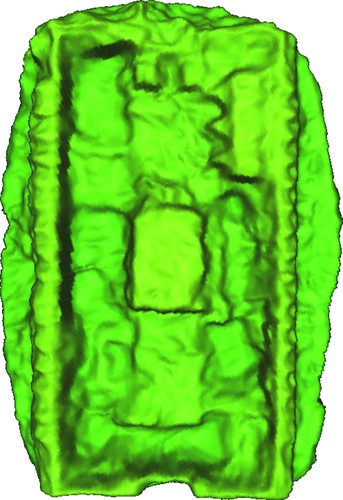}
 \\ \hline
  \rotatebox[origin=lc]{90}{ROI4}
    & \includegraphics[width=1.85cm,angle=90,origin=c]{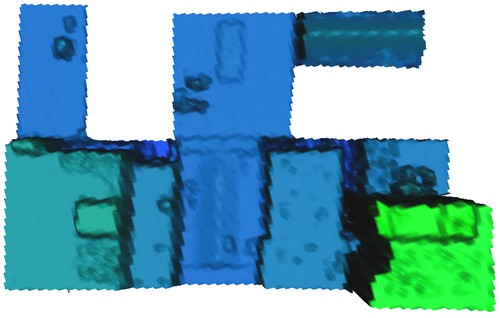}
    & \includegraphics[width=1.85cm,angle=90,origin=c]{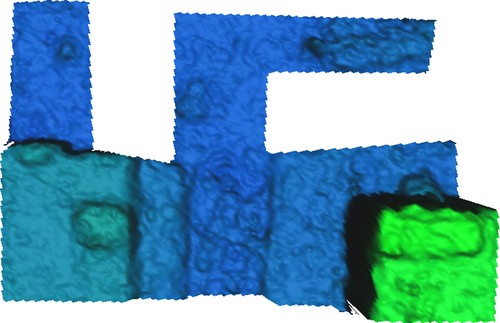}
    & \includegraphics[width=1.85cm,angle=90,origin=c]{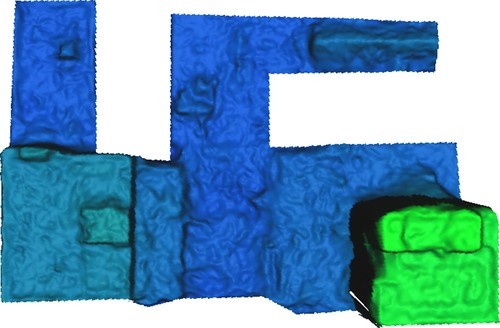}
    & \includegraphics[width=1.85cm,angle=90,origin=c]{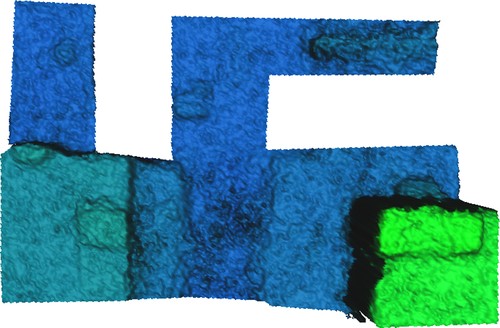}
    & \includegraphics[width=1.85cm,angle=90,origin=c]{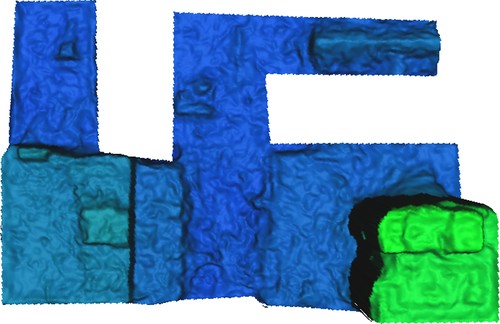}
    & \includegraphics[width=1.85cm,angle=90,origin=c]{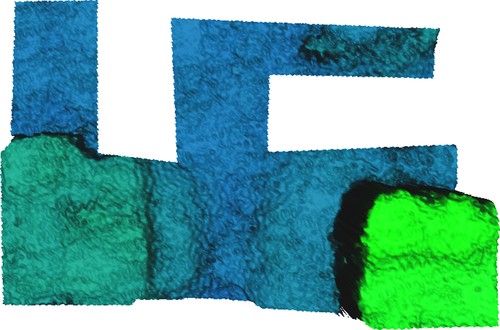}
    & \includegraphics[width=1.85cm,angle=90,origin=c]{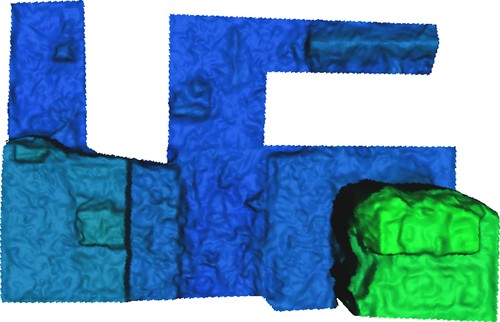}
 \\ \hline
    \end{tabular}
    \captionof{table}{Visualization of the building evaluation for the JAX test site from table \ref{table:evaluate_JAX_num}.}%
\label{table:evaluate_JAX}
\end{center}
\end{table}
\begin{table}
\begin{center}
\centering
\begin{tabular}{ @{}| c @{\hspace{1mm}}|@{\hspace{1mm}} c @{\hspace{1mm}}|@{\hspace{1mm}} c @{\hspace{1mm}}|@{\hspace{1mm}} c @{\hspace{1mm}}|@{\hspace{1mm}} c @{\hspace{1mm}}|@{\hspace{1mm}} c @{\hspace{1mm}}|@{\hspace{1mm}} c @{\hspace{1mm}}|@{\hspace{1mm}} c @{\hspace{1mm}}|@{}}
	\hline
	& LiDAR & tSGM & tSGM ref & S2P & SP2 ref & ASP & ASP ref \\ \hline
	  \rotatebox[origin=lc]{90}{ROI1}
    & \includegraphics[width=1.2cm,angle=90,origin=c]{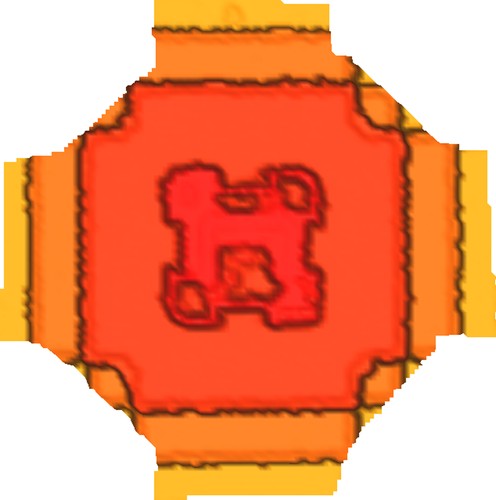}    
    & \includegraphics[width=1.2cm,angle=90,origin=c]{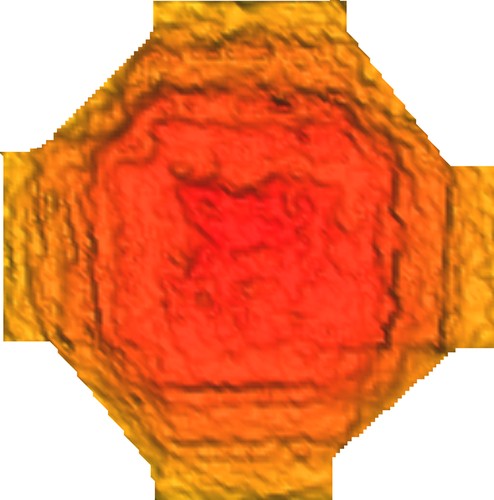}
    & \includegraphics[width=1.2cm,angle=90,origin=c]{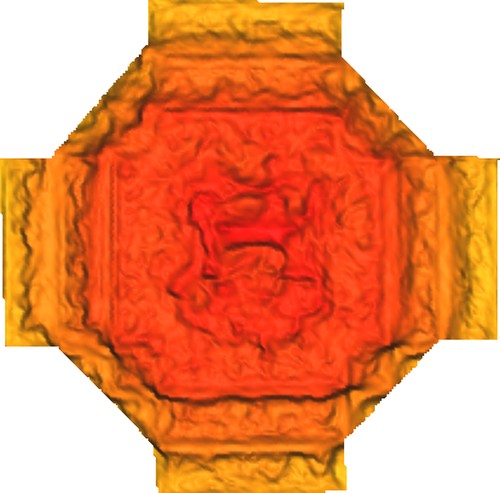}
    &\includegraphics[width=1.2cm,angle=90,origin=c]{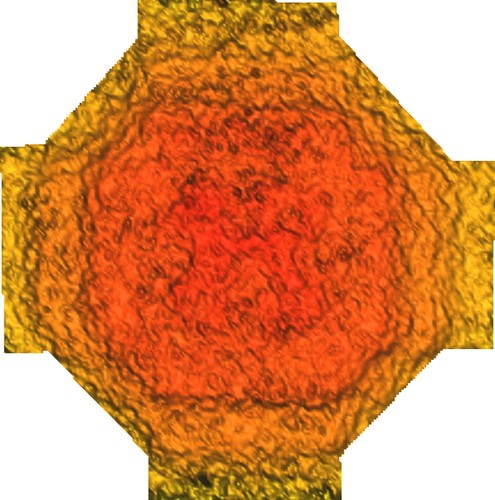}
    & \includegraphics[width=1.2cm,angle=90,origin=c]{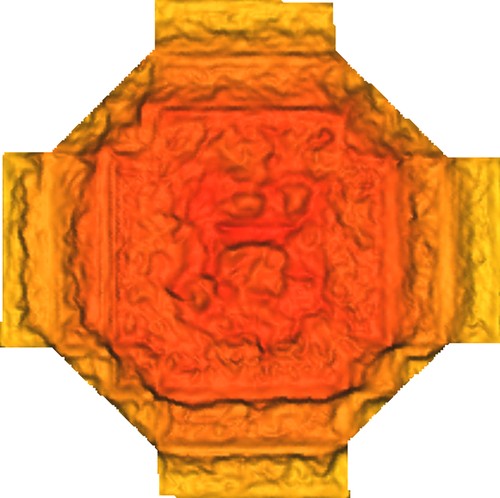}
    & \includegraphics[width=1.2cm,angle=90,origin=c]{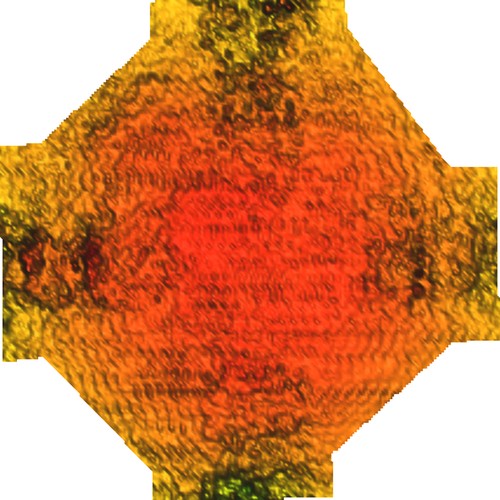}
    & \includegraphics[width=1.2cm,angle=90,origin=c]{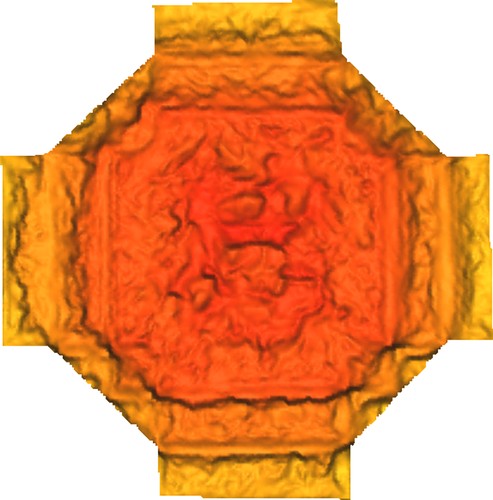}
 \\ \hline
  \rotatebox[origin=lc]{90}{ROI2}
    & \includegraphics[width=1.3cm]{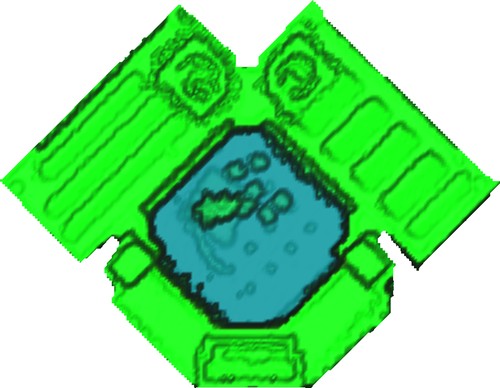}
    & \includegraphics[width=1.3cm]{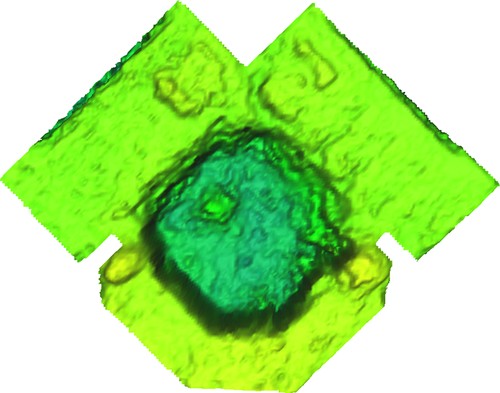}
    & \includegraphics[width=1.3cm]{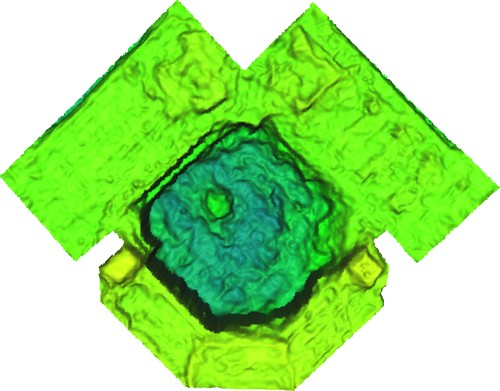}
    & \includegraphics[width=1.3cm]{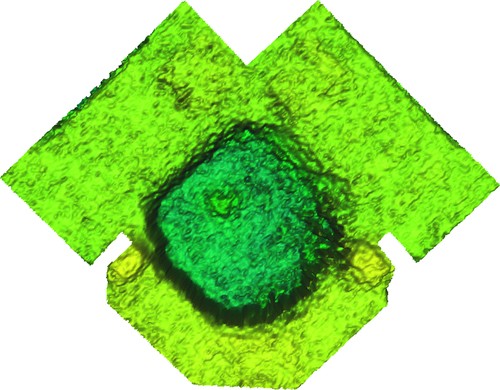}
    & \includegraphics[width=1.3cm]{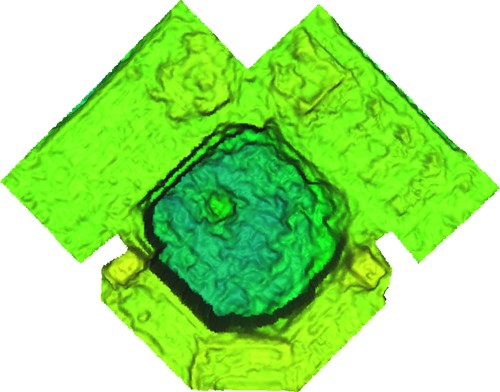}
    & \includegraphics[width=1.3cm]{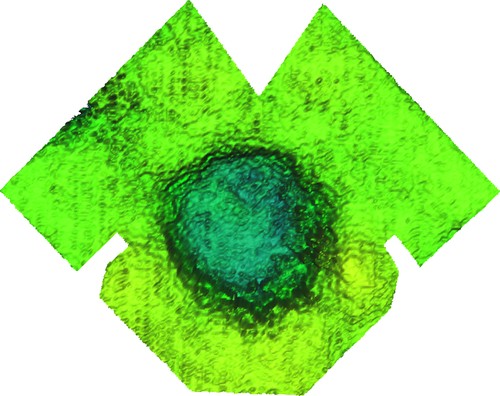}
    & \includegraphics[width=1.3cm]{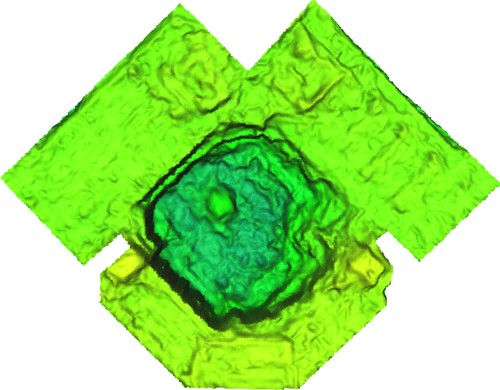}
 \\ \hline
  \rotatebox[origin=lc]{90}{ROI3}
     & \includegraphics[width=1.1cm]{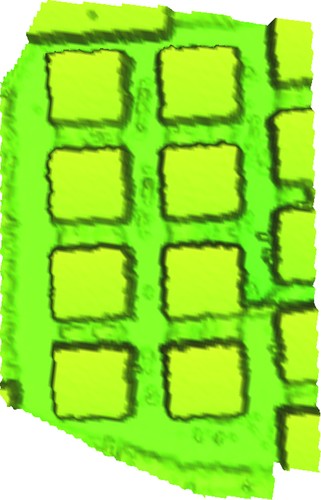}
    & \includegraphics[width=1.1cm]{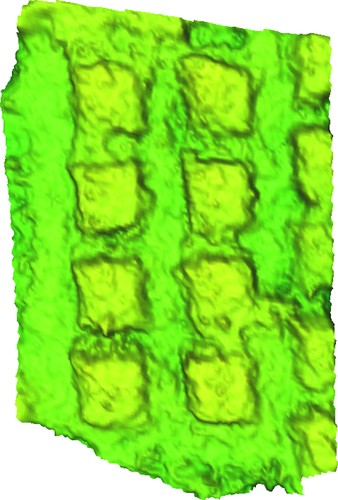}
    & \includegraphics[width=1.1cm]{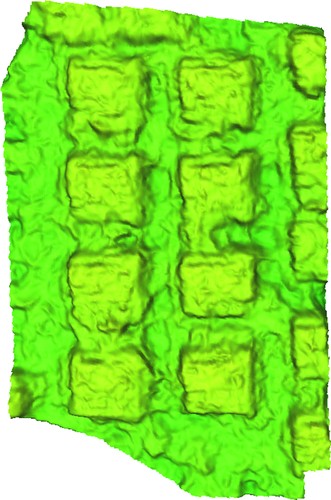}
    & \includegraphics[width=1.1cm]{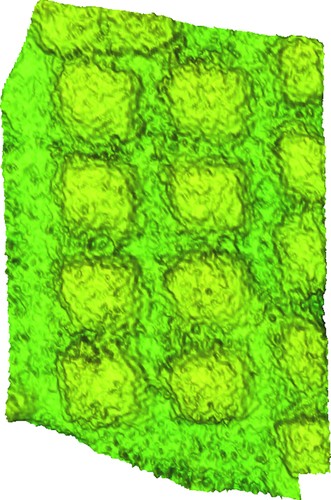}
    & \includegraphics[width=1.1cm]{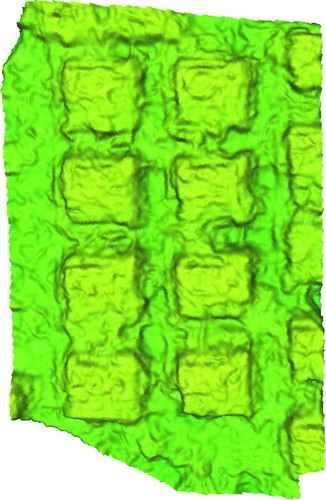}
    & \includegraphics[width=1.1cm]{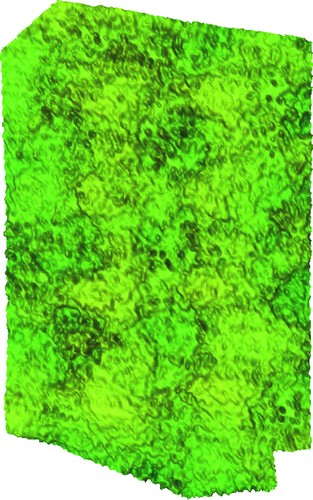}
    & \includegraphics[width=1.1cm]{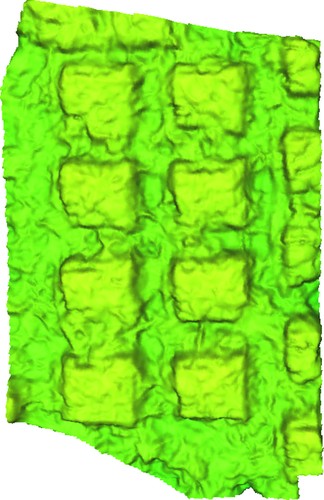}
 \\ \hline
  \rotatebox[origin=lc]{90}{ROI4}
      & \includegraphics[width=3.2cm,angle=90,origin=c]{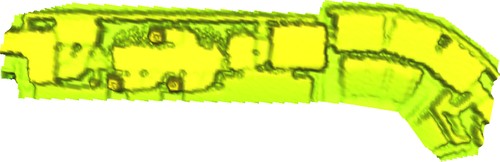}
    & \includegraphics[width=3.2cm,angle=90,origin=c]{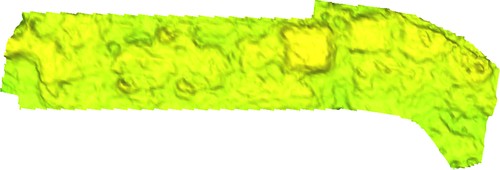}
    & \includegraphics[width=3.2cm,angle=90,origin=c]{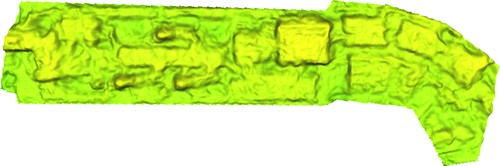}
    & \includegraphics[width=3.2cm,angle=90,origin=c]{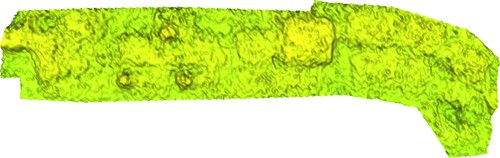}
    & \includegraphics[width=3.2cm,angle=90,origin=c]{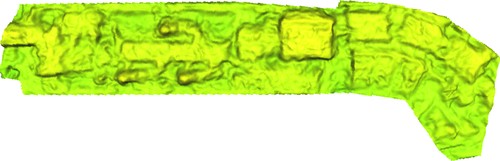}
    & \includegraphics[width=3.2cm,angle=90,origin=c]{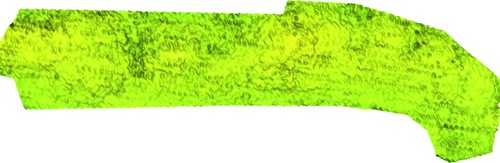}
    & \includegraphics[width=3.2cm,angle=90,origin=c]{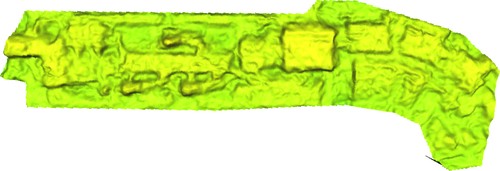}
 \\ \hline
    \end{tabular}
    \captionof{table}{Visualization of the building evaluation for the UCSD test site from table \ref{table:evaluate_UCSD_num}.} %
\label{table:evaluate_UCSD}
\end{center}
\end{table}
Figures \ref{table:evaluate_JAX} and \ref{table:evaluate_UCSD} display exemplary surface reconstructions of the different tested methods for both ROIs. Tables \ref{table:evaluate_JAX_num} and \ref{table:evaluate_UCSD_num} list the corresponding quantitative results. Again, mesh refinement generally improves the accuracy over SGM and ASP, at both sites. With S2P we get mixed results. Its accuracy is already very high (for several buildings at, or below GSD), so the results after refinement are quantitatively almost the same. Still, visual inspection makes clear that the quantitative metrics do not fully characterise the model quality:
The refined surfaces are crisper and more correct on complicated roof structures. The numbers do not reflect this, because after refinement the surface tends to be slightly noisier on flat, fronto-parallel areas that play to the strength of the constant-height prior built into most MVS and 2.5D fusion methods.
Overall, we observe crisper reconstructions of roof details after refinement. These details are recovered even when starting from the roughest initial surface (generated with NASA's ASP), where they are completely missing.
The meshes after refinement are also visually comparable, independent of the initialisation, indicating favourable convergence properties of the hierarchical optimisation.
We note in passing that the 68-percentile metric is comparable with the numbers published in \cite{brown2018large}.

\section{Conclusion}
We have presented an algorithm to refine 3D surface meshes by minimising the photometric transfer error between multiple pairs of satellite images, whose sensor poses are specified through the RFM.

In experiments on high-resolution data from WorldView-3, we have shown that, compared to the input from state-of-the-art MVS pipelines, the refinement 
extracts additional 3D surface details and better reproduces crisp edges and discontinuities. It also decreases the residual errors of the recovered surface in most cases. Being based on image similarity, it is however still challenged by repetitive texture and non-Lambertian surfaces.
To further diminish artefacts of our method in such areas, one important direction is to endow 3D surface models with more a-priori knowledge about the reconstructed scene. In particular, our current implementation includes only a simple thin-plate regulariser to favour low curvature. Elementary priors like the preference for piecewise constant heights (extensively used in dense matching and 2.5D model fusion), or a preference for right angles, are still missing and need to be ported to the world of 3D mesh-based reconstruction.

A major difference to existing reconstruction pipelines is that our algorithm is able to handle true 3D geometry. We have shown that, when including oblique satellite views, the method produces cleaner facade structures and recovers facade details.  This property could be advantageous for downstream building modelling, surface structuring and analysis based on geometric features, and of course visual display from off-nadir viewpoints.
We believe that with the growing availability of high-resolution, oblique satellite images the field could move beyond 2.5D elevation models and the
quest for full 3D models will gain momentum. We hope that our work triggers further research in that direction.

\clearpage
\bibliography{mybibfile}

\begin{thebibliography}{71}
\expandafter\ifx\csname natexlab\endcsname\relax\def\natexlab#1{#1}\fi
\providecommand{\url}[1]{\texttt{#1}}
\providecommand{\href}[2]{#2}
\providecommand{\path}[1]{#1}
\providecommand{\DOIprefix}{doi:}
\providecommand{\ArXivprefix}{arXiv:}
\providecommand{\URLprefix}{URL: }
\providecommand{\Pubmedprefix}{pmid:}
\providecommand{\doi}[1]{\href{http://dx.doi.org/#1}{\path{#1}}}
\providecommand{\Pubmed}[1]{\href{pmid:#1}{\path{#1}}}
\providecommand{\bibinfo}[2]{#2}
\ifx\xfnm\relax \def\xfnm[#1]{\unskip,\space#1}\fi
\bibitem[{Besse(2013)}]{besse2013patchmatch}
\bibinfo{author}{Besse, F.O.}, \bibinfo{year}{2013}.
\newblock \bibinfo{title}{PatchMatch Belief Propagation for Correspondence
  Field Estimation and its Applications}.
\newblock Ph.D. thesis. University College London (UCL).
\bibitem[{Blaha et~al.(2017)Blaha, Rothermel, Oswald, Sattler, Richard, Wegner,
  Pollefeys and Schindler}]{blaha2017semantically}
\bibinfo{author}{Blaha, M.}, \bibinfo{author}{Rothermel, M.},
  \bibinfo{author}{Oswald, M.R.}, \bibinfo{author}{Sattler, T.},
  \bibinfo{author}{Richard, A.}, \bibinfo{author}{Wegner, J.D.},
  \bibinfo{author}{Pollefeys, M.}, \bibinfo{author}{Schindler, K.},
  \bibinfo{year}{2017}.
\newblock \bibinfo{title}{Semantically informed multiview surface refinement},
  in: \bibinfo{booktitle}{Proceedings of the IEEE International Conference on
  Computer Vision}, pp. \bibinfo{pages}{3819--3827}.
\bibitem[{Bosch et~al.(2016)Bosch, Kurtz, Hagstrom and Brown}]{Bosch2016}
\bibinfo{author}{Bosch, M.}, \bibinfo{author}{Kurtz, Z.},
  \bibinfo{author}{Hagstrom, S.}, \bibinfo{author}{Brown, M.},
  \bibinfo{year}{2016}.
\newblock \bibinfo{title}{A multiple view stereo benchmark for satellite
  imagery}, in: \bibinfo{booktitle}{Proceedings of the IEEE Applied Imagery
  Pattern Recognition Workshop (AIPR)}, pp. \bibinfo{pages}{1--9}.
\bibitem[{Bradski(2000)}]{opencv_library}
\bibinfo{author}{Bradski, G.}, \bibinfo{year}{2000}.
\newblock \bibinfo{title}{{The OpenCV Library}}.
\newblock \bibinfo{journal}{Dr. Dobb's Journal of Software Tools} .
\bibitem[{Brown et~al.(2018)Brown, Goldberg, Foster, Leichtman, Wang, Hagstrom,
  Bosch and Almes}]{brown2018large}
\bibinfo{author}{Brown, M.}, \bibinfo{author}{Goldberg, H.},
  \bibinfo{author}{Foster, K.}, \bibinfo{author}{Leichtman, A.},
  \bibinfo{author}{Wang, S.}, \bibinfo{author}{Hagstrom, S.},
  \bibinfo{author}{Bosch, M.}, \bibinfo{author}{Almes, S.},
  \bibinfo{year}{2018}.
\newblock \bibinfo{title}{Large-scale public lidar and satellite image data set
  for urban semantic labeling}, in: \bibinfo{booktitle}{Proceedings of Laser
  Radar Technology and Applications XXIII}, pp. \bibinfo{pages}{154--167}.
\bibitem[{d'Angelo(2016)}]{d2016improving}
\bibinfo{author}{d'Angelo, P.}, \bibinfo{year}{2016}.
\newblock \bibinfo{title}{Improving semi-global matching: Cost aggregation and
  confidence measure}.
\newblock \bibinfo{journal}{International Archives of the Photogrammetry,
  Remote Sensing and Spatial Information Sciences} \bibinfo{volume}{41},
  \bibinfo{pages}{299--304}.
\bibitem[{d'Angelo and Kuschk(2012)}]{dAngelo2012}
\bibinfo{author}{d'Angelo, P.}, \bibinfo{author}{Kuschk, G.},
  \bibinfo{year}{2012}.
\newblock \bibinfo{title}{Dense multi-view stereo from satellite imagery}, in:
  \bibinfo{booktitle}{Proceedings of 2012 IEEE International Geoscience and
  Remote Sensing Symposium}, pp. \bibinfo{pages}{6944--6947}.
\bibitem[{De~Franchis et~al.(2014)De~Franchis, Meinhardt-Llopis, Michel, Morel
  and Facciolo}]{DeFranchis2014}
\bibinfo{author}{De~Franchis, C.}, \bibinfo{author}{Meinhardt-Llopis, E.},
  \bibinfo{author}{Michel, J.}, \bibinfo{author}{Morel, J.M.},
  \bibinfo{author}{Facciolo, G.}, \bibinfo{year}{2014}.
\newblock \bibinfo{title}{An automatic and modular stereo pipeline for
  pushbroom images}.
\newblock \bibinfo{journal}{ISPRS Annals of the Photogrammetry, Remote Sensing
  and Spatial Information Sciences} \bibinfo{volume}{2},
  \bibinfo{pages}{49--56}.
\bibitem[{Delaunoy and Prados(2011)}]{delaunoy2011gradient}
\bibinfo{author}{Delaunoy, A.}, \bibinfo{author}{Prados, E.},
  \bibinfo{year}{2011}.
\newblock \bibinfo{title}{Gradient flows for optimizing triangular mesh-based
  surfaces: Applications to 3d reconstruction problems dealing with
  visibility}.
\newblock \bibinfo{journal}{International journal of computer vision}
  \bibinfo{volume}{95}, \bibinfo{pages}{100--123}.
\bibitem[{Delaunoy et~al.(2008)Delaunoy, Prados, Pirac{\'e}s, Pons and
  Sturm}]{delaunoy2008minimizing}
\bibinfo{author}{Delaunoy, A.}, \bibinfo{author}{Prados, E.},
  \bibinfo{author}{Pirac{\'e}s, P.G.I.}, \bibinfo{author}{Pons, J.P.},
  \bibinfo{author}{Sturm, P.}, \bibinfo{year}{2008}.
\newblock \bibinfo{title}{Minimizing the multi-view stereo reprojection error
  for triangular surface meshes}, in: \bibinfo{booktitle}{Proceedings of BMVC
  2008-British Machine Vision Conference}, pp. \bibinfo{pages}{1--10}.
\bibitem[{Dial and Grodecki(2002)}]{dial2002block}
\bibinfo{author}{Dial, G.}, \bibinfo{author}{Grodecki, J.},
  \bibinfo{year}{2002}.
\newblock \bibinfo{title}{Block adjustment with rational polynomial camera
  models}, in: \bibinfo{booktitle}{Proceedings of ASPRS 2002 Conference,
  Washington, DC}, pp. \bibinfo{pages}{22--26}.
\bibitem[{Drory et~al.(2014)Drory, Haubold, Avidan and
  Hamprecht}]{drory2014semi}
\bibinfo{author}{Drory, A.}, \bibinfo{author}{Haubold, C.},
  \bibinfo{author}{Avidan, S.}, \bibinfo{author}{Hamprecht, F.A.},
  \bibinfo{year}{2014}.
\newblock \bibinfo{title}{Semi-global matching: a principled derivation in
  terms of message passing}, in: \bibinfo{booktitle}{Proceedings of German
  Conference on Pattern Recognition}, pp. \bibinfo{pages}{43--53}.
\bibitem[{d’Angelo and Reinartz(2011)}]{dAngelo2011}
\bibinfo{author}{d’Angelo, P.}, \bibinfo{author}{Reinartz, P.},
  \bibinfo{year}{2011}.
\newblock \bibinfo{title}{Semiglobal matching results on the isprs stereo
  matching benchmark}.
\newblock \bibinfo{journal}{International Archives of the Photogrammetry,
  Remote Sensing and Spatial Information Sciences} \bibinfo{volume}{38},
  \bibinfo{pages}{79--84}.
\bibitem[{Facciolo et~al.(2015)Facciolo, De~Franchis and
  Meinhardt}]{facciolo2015mgm}
\bibinfo{author}{Facciolo, G.}, \bibinfo{author}{De~Franchis, C.},
  \bibinfo{author}{Meinhardt, E.}, \bibinfo{year}{2015}.
\newblock \bibinfo{title}{Mgm: A significantly more global matching for
  stereovision}, in: \bibinfo{booktitle}{Proceedings of BMVC 2015-British
  Machine Vision Conference}, pp. \bibinfo{pages}{1--12}.
\bibitem[{Facciolo et~al.(2017)Facciolo, De~Franchis and
  Meinhardt-Llopis}]{Facciolo2017}
\bibinfo{author}{Facciolo, G.}, \bibinfo{author}{De~Franchis, C.},
  \bibinfo{author}{Meinhardt-Llopis, E.}, \bibinfo{year}{2017}.
\newblock \bibinfo{title}{Automatic 3d reconstruction from multi-date satellite
  images}, in: \bibinfo{booktitle}{Proceedings of IEEE Conference on Computer
  Vision and Pattern Recognition}, pp. \bibinfo{pages}{57--66}.
\bibitem[{Fraser et~al.(2006)Fraser, Dial and Grodecki}]{fraser2006}
\bibinfo{author}{Fraser, C.}, \bibinfo{author}{Dial, G.},
  \bibinfo{author}{Grodecki, J.}, \bibinfo{year}{2006}.
\newblock \bibinfo{title}{Sensor orientation via {RPC}s}.
\newblock \bibinfo{journal}{ISPRS Journal of Photogrammetry and Remote Sensing}
  \bibinfo{volume}{60}, \bibinfo{pages}{182--194}.
\bibitem[{Fraser and Yamakawa(2004)}]{fraser2004}
\bibinfo{author}{Fraser, C.}, \bibinfo{author}{Yamakawa, T.},
  \bibinfo{year}{2004}.
\newblock \bibinfo{title}{Insights into the affine model for high-resolution
  satellite sensor orientation}.
\newblock \bibinfo{journal}{ISPRS Journal of Photogrammetry and Remote Sensing}
  \bibinfo{volume}{58}, \bibinfo{pages}{275--288}.
\bibitem[{Fuhrmann and Gösele(2014)}]{fuhrmann2014floating}
\bibinfo{author}{Fuhrmann, S.}, \bibinfo{author}{Gösele, M.},
  \bibinfo{year}{2014}.
\newblock \bibinfo{title}{Floating scale surface reconstruction}.
\newblock \bibinfo{journal}{ACM Transactions on Graphics (ToG)}
  \bibinfo{volume}{33}, \bibinfo{pages}{46:1--46:11}.
\bibitem[{Furukawa and Ponce(2010)}]{Furu:2010:PMVS}
\bibinfo{author}{Furukawa, Y.}, \bibinfo{author}{Ponce, J.},
  \bibinfo{year}{2010}.
\newblock \bibinfo{title}{Accurate, dense, and robust multi-view stereopsis}.
\newblock \bibinfo{journal}{IEEE Transactions on Pattern Analysis and Machine
  Intelligence} \bibinfo{volume}{32}, \bibinfo{pages}{1362--1376}.
\bibitem[{Fusiello et~al.(2000)Fusiello, Trucco and
  Verri}]{fusiello2000compact}
\bibinfo{author}{Fusiello, A.}, \bibinfo{author}{Trucco, E.},
  \bibinfo{author}{Verri, A.}, \bibinfo{year}{2000}.
\newblock \bibinfo{title}{A compact algorithm for rectification of stereo
  pairs}.
\newblock \bibinfo{journal}{Machine Vision and Applications}
  \bibinfo{volume}{12}, \bibinfo{pages}{16--22}.
\bibitem[{Galliani et~al.(2016)Galliani, Lasinger and
  Schindler}]{galliani2016gipuma}
\bibinfo{author}{Galliani, S.}, \bibinfo{author}{Lasinger, K.},
  \bibinfo{author}{Schindler, K.}, \bibinfo{year}{2016}.
\newblock \bibinfo{title}{Gipuma: Massively parallel multi-view stereo
  reconstruction}.
\newblock \bibinfo{journal}{Publikationen der Deutschen Gesellschaft f{\"u}r
  Photogrammetrie, Fernerkundung und Geoinformation e. V} \bibinfo{volume}{25},
  \bibinfo{pages}{361--369}.
\bibitem[{Gehrig and Franke(2007)}]{gehrig2007improving}
\bibinfo{author}{Gehrig, S.K.}, \bibinfo{author}{Franke, U.},
  \bibinfo{year}{2007}.
\newblock \bibinfo{title}{Improving stereo sub-pixel accuracy for long range
  stereo}, in: \bibinfo{booktitle}{Proceedings of 2007 IEEE 11th International
  Conference on Computer Vision}, pp. \bibinfo{pages}{1--7}.
\bibitem[{Gong and Fritsch(2019)}]{Gong2019}
\bibinfo{author}{Gong, K.}, \bibinfo{author}{Fritsch, D.},
  \bibinfo{year}{2019}.
\newblock \bibinfo{title}{{DSM} generation from high resolution multi-view
  stereo satellite imagery}.
\newblock \bibinfo{journal}{Photogrammetric Engineering \& Remote Sensing}
  \bibinfo{volume}{85}, \bibinfo{pages}{379--387}.
\bibitem[{G\"{o}sele et~al.(2007)G\"{o}sele, Snavely, Curless, Hoppe and
  Seitz}]{goesele2007multi}
\bibinfo{author}{G\"{o}sele, M.}, \bibinfo{author}{Snavely, N.},
  \bibinfo{author}{Curless, B.}, \bibinfo{author}{Hoppe, H.},
  \bibinfo{author}{Seitz, S.M.}, \bibinfo{year}{2007}.
\newblock \bibinfo{title}{Multi-view stereo for community photo collections},
  in: \bibinfo{booktitle}{Proceedings of 2007 IEEE 11th International
  Conference on Computer Vision}, pp. \bibinfo{pages}{1--8}.
\bibitem[{Hartley and Saxena(1997)}]{hartley1997cubic}
\bibinfo{author}{Hartley, R.I.}, \bibinfo{author}{Saxena, T.},
  \bibinfo{year}{1997}.
\newblock \bibinfo{title}{The cubic rational polynomial camera model}, in:
  \bibinfo{booktitle}{Proceedings of the DARPA Image Understanding Workshop},
  pp. \bibinfo{pages}{649--653}.
\bibitem[{Helava(1988)}]{helava1988object}
\bibinfo{author}{Helava, U.}, \bibinfo{year}{1988}.
\newblock \bibinfo{title}{Object-space least-squares correlation}, in:
  \bibinfo{booktitle}{(ACSM and American Society for Photogrammety and Remote
  Sensing, Annual Convention, Saint Louis, MO, Mar. 14-18, 1988)
  Photogrammetric Engineering and Remote Sensing,}, pp.
  \bibinfo{pages}{711--714}.
\bibitem[{Hirschmüller(2008)}]{Hirschmuller2008}
\bibinfo{author}{Hirschmüller, H.}, \bibinfo{year}{2008}.
\newblock \bibinfo{title}{Stereo processing by semiglobal matching and mutual
  information}.
\newblock \bibinfo{journal}{IEEE Transactions on pattern analysis and machine
  intelligence} \bibinfo{volume}{30}, \bibinfo{pages}{328--341}.
\bibitem[{Jancosek and Pajdla(2011)}]{jancosek2011multi}
\bibinfo{author}{Jancosek, M.}, \bibinfo{author}{Pajdla, T.},
  \bibinfo{year}{2011}.
\newblock \bibinfo{title}{Multi-view reconstruction preserving weakly-supported
  surfaces}, in: \bibinfo{booktitle}{Proceedings of IEEE Conference on Computer
  Vision and Pattern Recognition}, pp. \bibinfo{pages}{3121--3128}.
\bibitem[{Kazhdan and Hoppe(2013)}]{kazhdan2013screened}
\bibinfo{author}{Kazhdan, M.}, \bibinfo{author}{Hoppe, H.},
  \bibinfo{year}{2013}.
\newblock \bibinfo{title}{Screened poisson surface reconstruction}.
\newblock \bibinfo{journal}{ACM Transactions on Graphics (ToG)}
  \bibinfo{volume}{32}, \bibinfo{pages}{29:1--29:13}.
\bibitem[{Kim(2000)}]{Kim2000}
\bibinfo{author}{Kim, T.}, \bibinfo{year}{2000}.
\newblock \bibinfo{title}{A study on the epipolarity of linear pushbroom
  images}.
\newblock \bibinfo{journal}{Photogrammetric Engineering \& Remote Sensing}
  \bibinfo{volume}{62}, \bibinfo{pages}{961--966}.
\bibitem[{Kim and Dowman(2006)}]{kim2006comparison}
\bibinfo{author}{Kim, T.}, \bibinfo{author}{Dowman, I.}, \bibinfo{year}{2006}.
\newblock \bibinfo{title}{Comparison of two physical sensor models for
  satellite images: position--rotation model and orbit--attitude model}.
\newblock \bibinfo{journal}{The Photogrammetric Record} \bibinfo{volume}{21},
  \bibinfo{pages}{110--123}.
\bibitem[{Knapitsch et~al.(2017)Knapitsch, Park, Zhou and
  Koltun}]{Knapitsch2017}
\bibinfo{author}{Knapitsch, A.}, \bibinfo{author}{Park, J.},
  \bibinfo{author}{Zhou, Q.Y.}, \bibinfo{author}{Koltun, V.},
  \bibinfo{year}{2017}.
\newblock \bibinfo{title}{Tanks and temples: Benchmarking large-scale scene
  reconstruction}.
\newblock \bibinfo{journal}{ACM Transactions on Graphics} \bibinfo{volume}{36},
  \bibinfo{pages}{78:1--78:13}.
\bibitem[{Kobbelt et~al.(1998)Kobbelt, Campagna, Vorsatz and Seidel}]{Kobbelt}
\bibinfo{author}{Kobbelt, L.}, \bibinfo{author}{Campagna, S.},
  \bibinfo{author}{Vorsatz, J.}, \bibinfo{author}{Seidel, H.P.},
  \bibinfo{year}{1998}.
\newblock \bibinfo{title}{Interactive multi-resolution modeling on arbitrary
  meshes}, in: \bibinfo{booktitle}{Proceedings of the 25th Annual Conference on
  Computer Graphics and Interactive Techniques}, pp. \bibinfo{pages}{105--114}.
\bibitem[{Kratky(1989)}]{kratky1989line}
\bibinfo{author}{Kratky, V.}, \bibinfo{year}{1989}.
\newblock \bibinfo{title}{On-line aspects of stereophotogrammetric processing
  of spot images}.
\newblock \bibinfo{journal}{Photogrammetric Engineering and Remote Sensing}
  \bibinfo{volume}{55}, \bibinfo{pages}{311--316}.
\bibitem[{Kuschk(2013)}]{Kuschk2013}
\bibinfo{author}{Kuschk, G.}, \bibinfo{year}{2013}.
\newblock \bibinfo{title}{Large scale urban reconstruction from remote sensing
  imagery}.
\newblock \bibinfo{journal}{International Archives of the Photogrammetry,
  Remote Sensing and Spatial Information Sciences} \bibinfo{volume}{40},
  \bibinfo{pages}{139–146}.
\bibitem[{Kuschk et~al.(2017)Kuschk, d’Angelo, Gaudrie, Reinartz and
  Cremers}]{kuschk2017spatially}
\bibinfo{author}{Kuschk, G.}, \bibinfo{author}{d’Angelo, P.},
  \bibinfo{author}{Gaudrie, D.}, \bibinfo{author}{Reinartz, P.},
  \bibinfo{author}{Cremers, D.}, \bibinfo{year}{2017}.
\newblock \bibinfo{title}{Spatially regularized fusion of multiresolution
  digital surface models}.
\newblock \bibinfo{journal}{IEEE Transactions on Geoscience and Remote Sensing}
  \bibinfo{volume}{55}, \bibinfo{pages}{1477--1488}.
\bibitem[{Labatut et~al.(2009)Labatut, Pons and Keriven}]{labatut2009robust}
\bibinfo{author}{Labatut, P.}, \bibinfo{author}{Pons, J.P.},
  \bibinfo{author}{Keriven, R.}, \bibinfo{year}{2009}.
\newblock \bibinfo{title}{Robust and efficient surface reconstruction from
  range data}.
\newblock \bibinfo{journal}{Computer graphics forum} \bibinfo{volume}{28},
  \bibinfo{pages}{2275--2290}.
\bibitem[{Li et~al.(2016)Li, Siu, Fang and Quan}]{li2016efficient}
\bibinfo{author}{Li, S.}, \bibinfo{author}{Siu, S.Y.}, \bibinfo{author}{Fang,
  T.}, \bibinfo{author}{Quan, L.}, \bibinfo{year}{2016}.
\newblock \bibinfo{title}{Efficient multi-view surface refinement with adaptive
  resolution control}, in: \bibinfo{booktitle}{Proceedings of European
  Conference on Computer Vision}, pp. \bibinfo{pages}{349--364}.
\bibitem[{Li et~al.(2015)Li, Wang, Zuo, Meng and Zhang}]{li2015detail}
\bibinfo{author}{Li, Z.}, \bibinfo{author}{Wang, K.}, \bibinfo{author}{Zuo,
  W.}, \bibinfo{author}{Meng, D.}, \bibinfo{author}{Zhang, L.},
  \bibinfo{year}{2015}.
\newblock \bibinfo{title}{Detail-preserving and content-aware variational
  multi-view stereo reconstruction}.
\newblock \bibinfo{journal}{IEEE Transactions on Image Processing}
  \bibinfo{volume}{25}, \bibinfo{pages}{864--877}.
\bibitem[{Loop and Zhang(1999)}]{loop1999computing}
\bibinfo{author}{Loop, C.}, \bibinfo{author}{Zhang, Z.}, \bibinfo{year}{1999}.
\newblock \bibinfo{title}{Computing rectifying homographies for stereo vision},
  in: \bibinfo{booktitle}{Proceedings of IEEE Conference on Computer Vision and
  Pattern Recognition}, pp. \bibinfo{pages}{125--131}.
\bibitem[{Moratto et~al.(2010)Moratto, Broxton, Beyer, Lundy and
  Husmann}]{nasa_asp}
\bibinfo{author}{Moratto, Z.M.}, \bibinfo{author}{Broxton, M.J.},
  \bibinfo{author}{Beyer, R.A.}, \bibinfo{author}{Lundy, M.},
  \bibinfo{author}{Husmann, K.}, \bibinfo{year}{2010}.
\newblock \bibinfo{title}{Ames stereo pipeline, nasa's open source automated
  stereogrammetry software}, in: \bibinfo{booktitle}{Proceedings of Lunar and
  Planetary Science Conference}, p. \bibinfo{pages}{2364}.
\bibitem[{Oh(2011)}]{Oh2011}
\bibinfo{author}{Oh, J.}, \bibinfo{year}{2011}.
\newblock \bibinfo{title}{Novel Approach to Epipolar Resampling of HRSI and
  Satellite Stereo Imagery-based Georeferencing of Aerial Images}.
\newblock Ph.D. thesis. The Ohio State University.
\bibitem[{Ozcanli et~al.(2015)Ozcanli, Dong, Mundy, Webb, Hammoud and
  Tom}]{Ozcanli2015}
\bibinfo{author}{Ozcanli, O.C.}, \bibinfo{author}{Dong, Y.},
  \bibinfo{author}{Mundy, J.L.}, \bibinfo{author}{Webb, H.},
  \bibinfo{author}{Hammoud, R.}, \bibinfo{author}{Tom, V.},
  \bibinfo{year}{2015}.
\newblock \bibinfo{title}{A comparison of stereo and multiview 3-d
  reconstruction using cross-sensor satellite imagery}, in:
  \bibinfo{booktitle}{Proceedings of IEEE Conference on Computer Vision and
  Pattern Recognition}, pp. \bibinfo{pages}{17--25}.
\bibitem[{Patil et~al.(2019)Patil, Comandur, Prakash and Kak}]{avibench}
\bibinfo{author}{Patil, S.}, \bibinfo{author}{Comandur, B.},
  \bibinfo{author}{Prakash, T.}, \bibinfo{author}{Kak, A.C.},
  \bibinfo{year}{2019}.
\newblock \bibinfo{title}{A new stereo benchmarking dataset for satellite
  images}.
\newblock \bibinfo{journal}{Computing Research Repository}
  \bibinfo{volume}{abs/1907.04404}.
\newblock \URLprefix \url{http://arxiv.org/abs/1907.04404}.
\bibitem[{Poli and Toutin(2012)}]{poli2012review}
\bibinfo{author}{Poli, D.}, \bibinfo{author}{Toutin, T.}, \bibinfo{year}{2012}.
\newblock \bibinfo{title}{Review of developments in geometric modelling for
  high resolution satellite pushbroom sensors}.
\newblock \bibinfo{journal}{The Photogrammetric Record} \bibinfo{volume}{27},
  \bibinfo{pages}{58--73}.
\bibitem[{Pollard and Mundy(2007)}]{Pollard2007}
\bibinfo{author}{Pollard, T.}, \bibinfo{author}{Mundy, J.L.},
  \bibinfo{year}{2007}.
\newblock \bibinfo{title}{Change detection in a 3-d world}, in:
  \bibinfo{booktitle}{Proceedings of IEEE Conference on Computer Vision and
  Pattern Recognition}, pp. \bibinfo{pages}{1--6}.
\bibitem[{Pollard et~al.(2010)Pollard, Eden, Mundy and Cooper}]{Pollard2010}
\bibinfo{author}{Pollard, T.B.}, \bibinfo{author}{Eden, I.},
  \bibinfo{author}{Mundy, J.L.}, \bibinfo{author}{Cooper, D.B.},
  \bibinfo{year}{2010}.
\newblock \bibinfo{title}{A volumetric approach to change detection in
  satellite images}.
\newblock \bibinfo{journal}{Photogrammetric Engineering \& Remote Sensing}
  \bibinfo{volume}{76}, \bibinfo{pages}{817--831}.
\bibitem[{Pollefeys et~al.(1999)Pollefeys, Koch and
  Van~Gool}]{pollefeys1999simple}
\bibinfo{author}{Pollefeys, M.}, \bibinfo{author}{Koch, R.},
  \bibinfo{author}{Van~Gool, L.}, \bibinfo{year}{1999}.
\newblock \bibinfo{title}{A simple and efficient rectification method for
  general motion}, in: \bibinfo{booktitle}{Proceedings of the IEEE
  International Conference on Computer Vision}, pp. \bibinfo{pages}{496--501}.
\bibitem[{Qin(2017)}]{Qin2017}
\bibinfo{author}{Qin, R.}, \bibinfo{year}{2017}.
\newblock \bibinfo{title}{Automated 3d recovery from very high resolution
  multi-view satellite images}.
\newblock \bibinfo{journal}{Computing Research Repository}
  \bibinfo{volume}{abs/1905.07475}.
\newblock \URLprefix \url{https://arxiv.org/abs/1905.07475}.
\bibitem[{Reinartz et~al.(2010)Reinartz, d'Angelo, Krau{\ss}, Poli, Jacobsen
  and Buyuksalih}]{reinartz2010benchmarking}
\bibinfo{author}{Reinartz, P.}, \bibinfo{author}{d'Angelo, P.},
  \bibinfo{author}{Krau{\ss}, T.}, \bibinfo{author}{Poli, D.},
  \bibinfo{author}{Jacobsen, K.}, \bibinfo{author}{Buyuksalih, G.},
  \bibinfo{year}{2010}.
\newblock \bibinfo{title}{Benchmarking and quality analysis of dem generated
  from high and very high resolution optical stereo satellite data}, in:
  \bibinfo{booktitle}{Proceedings of The 2010 Canadian Geomatics Conference and
  Symposium of Commission I, ISPRS Convergence in Geomatics – Shaping
  Canada’s Competitive Landscape}, pp. \bibinfo{pages}{1--6}.
\bibitem[{Romanoni et~al.(2017)Romanoni, Ciccone, Visin and
  Matteucci}]{romanoni2017multi}
\bibinfo{author}{Romanoni, A.}, \bibinfo{author}{Ciccone, M.},
  \bibinfo{author}{Visin, F.}, \bibinfo{author}{Matteucci, M.},
  \bibinfo{year}{2017}.
\newblock \bibinfo{title}{Multi-view stereo with single-view semantic mesh
  refinement}, in: \bibinfo{booktitle}{Proceedings of the IEEE International
  Conference on Computer Vision}, pp. \bibinfo{pages}{706--715}.
\bibitem[{Roth and Mayer(2019)}]{roth2019reduction}
\bibinfo{author}{Roth, L.}, \bibinfo{author}{Mayer, H.}, \bibinfo{year}{2019}.
\newblock \bibinfo{title}{Reduction of the fronto-parallel bias for
  wide-baseline semi-global matching}.
\newblock \bibinfo{journal}{ISPRS Annals of Photogrammetry, Remote Sensing and
  Spatial Information Sciences} \bibinfo{volume}{4}, \bibinfo{pages}{69--76}.
\bibitem[{Rothermel et~al.(2012)Rothermel, Wenzel, Fritsch and
  Haala}]{rothermel2012sure}
\bibinfo{author}{Rothermel, M.}, \bibinfo{author}{Wenzel, K.},
  \bibinfo{author}{Fritsch, D.}, \bibinfo{author}{Haala, N.},
  \bibinfo{year}{2012}.
\newblock \bibinfo{title}{{SURE}: Photogrammetric surface reconstruction from
  imagery}, in: \bibinfo{booktitle}{Proceedings of LowCost 3D Workshop Berlin},
  pp. \bibinfo{pages}{1--9}.
\bibitem[{Scharstein et~al.(2017)Scharstein, Taniai and
  Sinha}]{scharstein2017semi}
\bibinfo{author}{Scharstein, D.}, \bibinfo{author}{Taniai, T.},
  \bibinfo{author}{Sinha, S.N.}, \bibinfo{year}{2017}.
\newblock \bibinfo{title}{Semi-global stereo matching with surface orientation
  priors}, in: \bibinfo{booktitle}{Proceedings of International Conference on
  3D Vision (3DV)}, pp. \bibinfo{pages}{215--224}.
\bibitem[{Sch\"{o}nberger et~al.(2016)Sch\"{o}nberger, Zheng, Pollefeys and
  Frahm}]{schoenberger2016mvs}
\bibinfo{author}{Sch\"{o}nberger, J.L.}, \bibinfo{author}{Zheng, E.},
  \bibinfo{author}{Pollefeys, M.}, \bibinfo{author}{Frahm, J.M.},
  \bibinfo{year}{2016}.
\newblock \bibinfo{title}{Pixelwise view selection for unstructured multi-view
  stereo}, in: \bibinfo{booktitle}{Proceedings of European Conference on
  Computer Vision}, pp. \bibinfo{pages}{501--518}.
\bibitem[{Sch\"ops et~al.(2017)Sch\"ops, Sch\"onberger, Galliani, Sattler,
  Schindler, Pollefeys and Geiger}]{schoeps2017cvpr}
\bibinfo{author}{Sch\"ops, T.}, \bibinfo{author}{Sch\"onberger, J.L.},
  \bibinfo{author}{Galliani, S.}, \bibinfo{author}{Sattler, T.},
  \bibinfo{author}{Schindler, K.}, \bibinfo{author}{Pollefeys, M.},
  \bibinfo{author}{Geiger, A.}, \bibinfo{year}{2017}.
\newblock \bibinfo{title}{A multi-view stereo benchmark with high-resolution
  images and multi-camera videos}, in: \bibinfo{booktitle}{Proceedings of IEEE
  conference on Computer Vision and Pattern Recognition}, pp.
  \bibinfo{pages}{3260--3269}.
\bibitem[{Seitz et~al.(2006)Seitz, Curless, Diebel, Scharstein and
  Szeliski}]{seitz2006comparison}
\bibinfo{author}{Seitz, S.M.}, \bibinfo{author}{Curless, B.},
  \bibinfo{author}{Diebel, J.}, \bibinfo{author}{Scharstein, D.},
  \bibinfo{author}{Szeliski, R.}, \bibinfo{year}{2006}.
\newblock \bibinfo{title}{A comparison and evaluation of multi-view stereo
  reconstruction algorithms}, in: \bibinfo{booktitle}{Proceedings of IEEE
  Conference on Computer Vision and Pattern Recognition}, pp.
  \bibinfo{pages}{519--528}.
\bibitem[{Shean et~al.(2016)Shean, Alexandrov, Moratto, Smith, Joughin, Porter
  and Morin}]{shean2016automated}
\bibinfo{author}{Shean, D.E.}, \bibinfo{author}{Alexandrov, O.},
  \bibinfo{author}{Moratto, Z.M.}, \bibinfo{author}{Smith, B.E.},
  \bibinfo{author}{Joughin, I.R.}, \bibinfo{author}{Porter, C.},
  \bibinfo{author}{Morin, P.}, \bibinfo{year}{2016}.
\newblock \bibinfo{title}{An automated, open-source pipeline for mass
  production of digital elevation models (dems) from very-high-resolution
  commercial stereo satellite imagery}.
\newblock \bibinfo{journal}{ISPRS Journal of Photogrammetry and Remote Sensing}
  \bibinfo{volume}{116}, \bibinfo{pages}{101--117}.
\bibitem[{Shimizu and Okutomi(2002)}]{shimizu2002precise}
\bibinfo{author}{Shimizu, M.}, \bibinfo{author}{Okutomi, M.},
  \bibinfo{year}{2002}.
\newblock \bibinfo{title}{Precise subpixel estimation on area-based matching}.
\newblock \bibinfo{journal}{Systems and Computers in Japan}
  \bibinfo{volume}{33}, \bibinfo{pages}{1--10}.
\bibitem[{Solem and Overgaard(2005)}]{solem2005geometric}
\bibinfo{author}{Solem, J.E.}, \bibinfo{author}{Overgaard, N.C.},
  \bibinfo{year}{2005}.
\newblock \bibinfo{title}{A geometric formulation of gradient descent for
  variational problems with moving surfaces}, in:
  \bibinfo{booktitle}{Proceedings of International Conference on Scale-Space
  and PDE Methods in Computer Vision}, pp. \bibinfo{pages}{419--430}.
\bibitem[{Szeliski and Scharstein(2004)}]{szeliski2004sampling}
\bibinfo{author}{Szeliski, R.}, \bibinfo{author}{Scharstein, D.},
  \bibinfo{year}{2004}.
\newblock \bibinfo{title}{Sampling the disparity space image}.
\newblock \bibinfo{journal}{IEEE Transactions on Pattern Analysis and Machine
  Intelligence} \bibinfo{volume}{26}, \bibinfo{pages}{419--425}.
\bibitem[{Tao and Hu(2001)}]{tao2001comprehensive}
\bibinfo{author}{Tao, C.V.}, \bibinfo{author}{Hu, Y.}, \bibinfo{year}{2001}.
\newblock \bibinfo{title}{A comprehensive study of the rational function model
  for photogrammetric processing}.
\newblock \bibinfo{journal}{Photogrammetric engineering and remote sensing}
  \bibinfo{volume}{67}, \bibinfo{pages}{1347--1358}.
\bibitem[{Ummenhofer and Brox(2015)}]{ummenhofer2015global}
\bibinfo{author}{Ummenhofer, B.}, \bibinfo{author}{Brox, T.},
  \bibinfo{year}{2015}.
\newblock \bibinfo{title}{Global, dense multiscale reconstruction for a billion
  points}, in: \bibinfo{booktitle}{Proceedings of the IEEE International
  Conference on Computer Vision}, pp. \bibinfo{pages}{1341--1349}.
\bibitem[{Vu et~al.(2012)Vu, Labatut, Pons and Keriven}]{vu2012high}
\bibinfo{author}{Vu, H.H.}, \bibinfo{author}{Labatut, P.},
  \bibinfo{author}{Pons, J.P.}, \bibinfo{author}{Keriven, R.},
  \bibinfo{year}{2012}.
\newblock \bibinfo{title}{High accuracy and visibility-consistent dense
  multiview stereo}.
\newblock \bibinfo{journal}{IEEE Transactions on Pattern Analysis and Machine
  Intelligence} \bibinfo{volume}{34}, \bibinfo{pages}{889--901}.
\bibitem[{Wang and Frahm(2017)}]{Wang2017}
\bibinfo{author}{Wang, K.}, \bibinfo{author}{Frahm, J.M.},
  \bibinfo{year}{2017}.
\newblock \bibinfo{title}{Fast and accurate satellite multi-view stereo using
  edge-aware interpolation}, in: \bibinfo{booktitle}{Proceedings of
  International Conference on 3D Vision (3DV)}, pp. \bibinfo{pages}{365--373}.
\bibitem[{Wang et~al.(2016)Wang, Stutts, Dunn and Frahm}]{Wang2016}
\bibinfo{author}{Wang, K.}, \bibinfo{author}{Stutts, C.},
  \bibinfo{author}{Dunn, E.}, \bibinfo{author}{Frahm, J.M.},
  \bibinfo{year}{2016}.
\newblock \bibinfo{title}{Efficient joint stereo estimation and land usage
  classification for multiview satellite data}, in:
  \bibinfo{booktitle}{Proceedings of IEEE Winter Conference on Applications of
  Computer Vision (WACV)}, pp. \bibinfo{pages}{1--9}.
\bibitem[{Wang et~al.(2010)Wang, Hu and Li}]{Wang2010}
\bibinfo{author}{Wang, M.}, \bibinfo{author}{Hu, F.}, \bibinfo{author}{Li, J.},
  \bibinfo{year}{2010}.
\newblock \bibinfo{title}{Epipolar arrangement of satellite imagery by
  projection trajectory simplification}.
\newblock \bibinfo{journal}{The Photogrammetric Record} \bibinfo{volume}{25},
  \bibinfo{pages}{422--436}.
\bibitem[{Wang et~al.(2011)Wang, Hu and Li}]{Wang2011}
\bibinfo{author}{Wang, M.}, \bibinfo{author}{Hu, F.}, \bibinfo{author}{Li, J.},
  \bibinfo{year}{2011}.
\newblock \bibinfo{title}{Epipolar resampling of linear pushbroom satellite
  imagery by a new epipolarity model}.
\newblock \bibinfo{journal}{ISPRS Journal of Photogrammetry and Remote Sensing}
  \bibinfo{volume}{66}, \bibinfo{pages}{347--355}.
\bibitem[{Wohlfeil et~al.(2012)Wohlfeil, Hirschmüller, Piltz, Börner and
  Suppa}]{Wohlfeil2012}
\bibinfo{author}{Wohlfeil, J.}, \bibinfo{author}{Hirschmüller, H.},
  \bibinfo{author}{Piltz, B.}, \bibinfo{author}{Börner, A.},
  \bibinfo{author}{Suppa, M.}, \bibinfo{year}{2012}.
\newblock \bibinfo{title}{Fully automated generation of accurate digital
  surface models with sub-meter resolution from satellite imagery}.
\newblock \bibinfo{journal}{International Archives of Photogrammetry, Remote
  Sensing and Spatial Information Sciences} \bibinfo{volume}{39},
  \bibinfo{pages}{75--80}.
\bibitem[{Wrobel(1987)}]{wrobel1987facets}
\bibinfo{author}{Wrobel, B.}, \bibinfo{year}{1987}.
\newblock \bibinfo{title}{Facets stereo vision (fast vision)—a new approach
  to computer stereo vision and to digital photogrammetry}, in:
  \bibinfo{booktitle}{ISPRS Intercommission Conf. Fast Processing of
  Photogrammetric Data}, pp. \bibinfo{pages}{231--258}.
\bibitem[{Zach et~al.(2007)Zach, Pock and Bischof}]{zach2007globally}
\bibinfo{author}{Zach, C.}, \bibinfo{author}{Pock, T.},
  \bibinfo{author}{Bischof, H.}, \bibinfo{year}{2007}.
\newblock \bibinfo{title}{A globally optimal algorithm for robust tv-l 1 range
  image integration}, in: \bibinfo{booktitle}{Proceedings of IEEE International
  Conference on Computer Vision}, pp. \bibinfo{pages}{1--8}.

\end{thebibliography}

\end{document}